\documentclass[journal]{IEEEtran}

\usepackage{multirow}

\ifCLASSINFOpdf
  \usepackage[pdftex]{graphicx}
\else
\fi

\usepackage{amsmath}

\usepackage{algorithmic}

\usepackage{array}

\usepackage{cite}
\usepackage{graphicx}
\usepackage{bm}
\usepackage{caption}
\usepackage{hyperref}
\usepackage{amssymb}
\usepackage{color,xcolor}
\usepackage{subfigure}

\usepackage{ragged2e}

\begin{document}
%
\title{Analyzing Unaligned Multimodal Sequence via Graph Convolution and Graph Pooling Fusion}
%
%
%

\author{Sijie Mai,
        Songlong Xing,
        Jiaxuan He,
        Ying Zeng,
        Haifeng Hu
\IEEEcompsocitemizethanks{\IEEEcompsocthanksitem Haifeng Hu (corresponding author)
 is with the School of Electronics and Information Technology, Sun Yat-sen University,
China.\protect\\
E-mail: huhaif@mail.sysu.edu.cn
\IEEEcompsocthanksitem Sijie Mai, Songlong Xing, Jiaxuan He, and Ying Zeng are with the School of Electronics and Information Technology, Sun Yat-sen University,
China.
\IEEEcompsocthanksitem This work was supported in part by the National Natural Science
Foundation of China under Grants 61673402, in part by the Natural Science Foundation of Guangdong under Grants 2017A030311029, in part by the Science and Technology Program of Guangzhou under Grants 201704020180.
}
\thanks{ }}


\IEEEtitleabstractindextext{%
\begin{abstract}
\justifying
In this paper, we study the task of multimodal sequence analysis which aims to draw inferences from visual, language and acoustic sequences. A majority of existing works generally focus on aligned fusion, mostly at word level, of the three modalities to accomplish this task, which is impractical in real-world scenarios. To overcome this issue, we seek to address the task of multimodal sequence analysis on unaligned modality sequences which is still relatively underexplored and also more challenging. Recurrent neural network (RNN) and its variants are widely used in multimodal sequence analysis, but they are susceptible to the issues of gradient vanishing/explosion and high time complexity due to its recurrent nature. Therefore, we propose a novel model, termed Multimodal Graph, to investigate the effectiveness of graph neural networks (GNN) on modeling multimodal sequential data. The graph-based structure enables parallel computation in time dimension and can learn longer temporal dependency in long unaligned sequences. Specifically, our Multimodal Graph is hierarchically structured to cater to two stages, i.e., intra- and inter-modal dynamics learning. For the first stage, a graph convolutional network is employed for each modality to learn intra-modal dynamics. In the second stage, given that the multimodal sequences are unaligned, the commonly considered word-level fusion does not pertain. To this end, we devise a graph pooling fusion network to automatically learn the associations between various nodes from different modalities. Additionally, we define multiple ways to construct the adjacency matrix for sequential data. Experimental results suggest that our graph-based model reaches state-of-the-art performance on two benchmark datasets.

\end{abstract}

}

\maketitle

\IEEEdisplaynontitleabstractindextext

%
\IEEEpeerreviewmaketitle

\section{\textbf{Introduction}}\label{sec:Introduction}
\IEEEPARstart{W}ith the development of the Internet and social platforms, there has been a large number of videos produced by users to express their views and posted online, which provides a source of multimodal data to analyze people's opinion in large quantities. As video is a typical form of multimodal information, people do not rely only on the spoken language, but they also use facial expressions, gestures and acoustic tones to convey information \cite{RMFN}. In this paper, our downstream task is to use three modalities, i,e., language, visual and acoustic modalities, to draw inferences for the sentiment polarities of speakers. These three modalities are complementary and actively interact with one another, providing more comprehensive information than one single modality. Hence, to maximally utilize the various information sources to capture the speaker's opinion with a multimodal architecture is a heated topic in multimodal sequence (language) analysis.



In the task of multimodal sequence analysis, two fundamental challenges exist, i.e., to learn the intra-modal dynamics of each modality, and the inter-modal counterpart for capturing cross-modal interactions \cite{Zadeh2017Tensor}. Previous researches mostly employ recurrent neural network (RNN) and its variants \cite{RNN,Hochreiter1997Long,Cho2014Learning} to learn these two aspects \cite{Multilogue-Net,RAVEN,RMFN,Zadeh2018Memory}, which, however, are slow in the inference process due to their recurrence in the time dimension. They are also prone to the problems of gradient vanishing/exploding and have limited capacity of learning long-term dependency \cite{Bengio1994Learning}, which adds to the difficulty in applying to the learning of intra- and inter-modal dynamics. Particularly, it is of great significance to learn longer temporal dependency for unaligned multimodal sequence analysis because the unaligned sequences are often very long.

Exploring effective approaches to cross-modal fusion to learn inter-modal dynamics has been one primary focus in the research of multimodal sequence analysis. To this end, a large portion of previous works fuse the three modalities at word level \cite{MFRM,RMFN,Zadeh2018Memory,mosei,RAVEN}. However, the interactions between various modalities are usually more complicated and last for longer than one word, i.e., cross-modal interactions may take place among words and the word-level fusion may break a complete interaction into multiple parts. Additionally, word-level fusion requires that the multimodal sequences are strictly aligned at word level. However, in real-world scenarios, the sequences are usually not aligned, for such alignment is time-consuming and computationally expensive \cite{P2FA}. Therefore, we claim that a fusion strategy should be able to dynamically associate various time steps from multiple modalities instead of considering cross-modal interaction at each specific time step.


To address the first issue, we propose to employ the high expressiveness of graph convolutional networks (GCNs) to model unimodal sequential signals as an alternative to RNN. Recently, GCNs have attracted significant attention to model graph structured data, and yielded state-of-the-art performance on a broad variety of tasks \cite{DIN, Micheli2009Neural, 4700287}. GCN demonstrates high effectiveness in learning the relevance of nodes via the operation of convolution, and importantly, they dispense with the need for recurrence and can be computed in parallel, which greatly boosts efficiency in inferring time compared to RNN. Existing researches on GCN mainly utilize it for modeling graph-structured data,
In contrast, in this work, we extend GCN to model sequential data and comparative analysis is conducted to show that GCN exhibits greater effectiveness compared to RNN \cite{DIN, Micheli2009Neural, 4700287} and temporal convolutional network (TCN) variants \cite{TCN,Trellis}. Specifically, we apply a GCN, i.e., GraphSAGE with mean pooling \cite{GraphSAGE}, for each modality to learn intra-modal dynamics. With the operation of graph convolution, longer temporal dependency can also be learnt by viewing each time step as a node and associating the relevant nodes even though they are far apart in time dimension.


One major obstacle to applying graph convolution to sequential learning is that sequences have no adjacency matrices as graph-structured data conventionally do. Therefore, the definition and deduction of adjacency matrices is important. In this paper, we present multiple ways to achieve this goal, including non-parametric methods and learnable methods. In the non-parametric way, we mainly investigate the effectiveness of a proposed matrix, namely generalized diagonal matrix, which is extremely fast and almost free of computation. In the learnable way, we automatically learn the adjacency matrix via gradient descent or node similarity (cross-node attention), which is more powerful and expressive. We present the comparative results of the proposed adjacency matrices in the experiment section.


For addressing the second problem, i.e., inter-modal dynamics learning, we elaborately design a graph pooling fusion network (GPFN), which learns to aggregate various nodes from different modalities by graph pooling. Firstly we analyze the rationality of mean/max graph pooling approaches via mathematical deduction. However, mean/max graph pooling are still subject to some limitations in that they are not learnable and can only fuse neighboring nodes. Hence, to fuse the nodes in a more expressive way, we further propose a pooling strategy, termed link similarity pooling, which, in comparison to mean/max pooling, is able to consider the association scores to the common neighbors of each two nodes. This learnable approach is based on the following two assumptions, i.e., (i) two nodes are similar or closely related if their neighbors significantly overlap and thus can be fused; and (ii) provided the two nodes are neighbors, they are integrable with a high possibility. The link similarity pooling thus automatically updates the adjacency matrix and node embeddings for learning a high-level graph representation.


In conclusion, we propose a brand new architecture named Multimodal Graph to address the sequential learning problem, particularly for unaligned multimodal sequence analysis. The contributions are listed below:
\begin{itemize}
  \item We propose a novel graph-based architecture to model multimodal sequences. Specifically, we design three unimodal GCNs to explore intra-modal dynamics of three modalities, and a graph pooling fusion network (GPFN) to explore cross-modal interactions and fuse various nodes from different modalities.
  \item In GPFN, we investigate mean/max pooling and link similarity pooling to cluster the nodes hierarchically and thus learn high-level graph representations.
  \item We introduce multiple approaches to define adjacency matrices for sequential data. And we compare the performance of different kinds of adjacency matrices empirically. Visualization of the adjacency matrices is also provided to give insight on multimodal sequence analysis.
  \item We show that the Multimodal Graph outperforms other methods on two widely-used benchmarks. Besides, the contrastive experiments against RNN and TCN variants demonstrate the effectiveness of GCN on modeling sequence, which indicates a novel approach in the research of sequence modeling tasks.
\end{itemize}


\section{\textbf{Related Work}}\label{sec:Related Work}
\label{sec:format}
\subsection{Multimodal Sequence Analysis}
Multimodal sequence analysis falls into the intersection of multimodal machine learning \cite{8269806} and multi-view learning \cite{Zadeh2018Memory, Multi-view, Xu2013A}, which has attracted significant research interest in recent years. Previous works focus on learning various fusion strategies to explore inter-modal (cross-view) dynamics. One of the simplest ways to explore inter-modal dynamics is to concatenate features at input feature level, which shows improvement over single modality \cite{Wollmer2013YouTube,Rozgic2012Ensemble,Poria2017Convolutional,Poria2017Context}. In contrast, a large number of publications firstly infer decision according to each modality and combine the decisions from all modalities using some voting mechanisms \cite{Wu2010Emotion,Nojavanasghari2016Deep,Personality,Zadeh2016MOSI}. However, as elaborated by Zadeh et al. \cite{Zadeh2017Tensor}, these two types of methods cannot effectively model inter-modal dynamics.

Consequently, more advanced fusion strategies are proposed in the past few years. Specifically, performing \textbf{tensor-based fusion} has received much attention \cite{T2FN, Liu2018Efficient,LMFN}. Tensor Fusion Network (TFN) \cite{Zadeh2017Tensor} and Low-rank Modality Fusion (LMF) \cite{Liu2018Efficient} adopt outer product to learn joint representation of three modalities. More recently, Mai et al. \cite{HFFN} propose a `Divide, Conquer and Combine' strategy to conduct local and global fusions. Similarly, Hierarchical Polynomial Fusion Network (HPFN) \cite{HPFN} is established to recursively integrate and transmit the local correlations into global correlations by multilinear fusion. Furthermore, some \textbf{translation methods} such as Multimodal Cyclic Translation Network (MCTN) \cite{MCTN} and Multimodal Transformer  \cite{mult} aim at learning a joint representation by translating source modality into target modality.
Mai et al. \cite{ARGF} propose to use adversarial training to translate distributions of different modalities and fuse the modalities using a graph fusion network which regards each modality as one node.
In comparison, our Multimodal Graph regards each time step as one node, using graph convolution and graph pooling to aggregate information and fuse nodes, respectively.
In addition, some approaches \cite{MFM,MMB} use \textbf{factorization methods} to learn inter-modal dynamics. Some methods such as Context-aware Interactive Attention (CIA) \cite{CIA}, Multi-Task Learning (MTL) \cite{MTL} and Multilogue-Net \cite{Multilogue-Net} that apply cross-modal attention mechanism to explore inter-modal dynamics are also very popular.

To avoid sarcasm and ambiguity, the mainstream in multimodal language analysis is to learn cross-view interactions at word level such that various modalities are aligned at time dimension \cite{Chen2018Multimodal,Gu2018Multimodal,RMFN,DHF,MFRM}. For instance, Memory Fusion Network (MFN) \cite{Zadeh2018Memory} uses systems of LSTM to learn intra-modal dynamics, and it implements delta-memory attention and multi-view gated memory network to fuse memories of LSTMs and explore inter-modal dynamics across time. Graph-MFN \cite{mosei} extends MFN by using a simple dynamic fusion graph to fuse features. Although Graph-MFN employs a graph network, it has major differences from our model.
Firstly, the dynamic fusion graph does not fuse features across time and merely regards each modality as one node and fuses features at the same time period, whereas we view each time step as one node and perform graph convolution over the node embeddings.
Secondly, we employ two operations of graph convolution and graph pooling for cross-modal fusion which are not involved in Graph-MFN.
Multi-attention Recurrent Network (MARN) \cite{Zadeh2018Multi} is developed for human communication comprehension to discover inter-modal dynamics over time using multi-attention block. In addition, Recurrent Attended Variation Embedding Network (RAVEN) \cite{RAVEN} models multimodal language by shifting word representations based on the features of audio and vision clues. However, in unaligned multimodal sequence, which is more practical in real-world scenarios and also more challenging, word-level fusion cannot be performed.

A clear distinction of all these previous methods between our Multimodal Graph is that we do not apply any RNN or TCN variants to learn intra-modal and inter-modal dynamics, but instead, we investigate the effectiveness of GCN on exploring multimodal sequence signal. Our Multimodal Graph is very elegant and effective, which can effectively learn longer temporal dependency by directly associating the distant related nodes and allow parallel computing at time dimension. 

\subsection{Graph Neural Networks}
A wide variety of graph neural networks (GNNs) have been proposed in the last few years \cite{DIN, Micheli2009Neural, 4700287}. We mainly focus on graph convolution networks (GCNs) \cite{GCN} in this paper. GCNs have become increasingly popular recently due to its applicability to graph structured data and yielded state-of-the-art performance on a variety of learning tasks, such as node classification \cite{GraphSAGE}, link prediction \cite{SEAL}, and graph classification \cite{DGCNN}.

Among all these GCN models, those trying to learn or recover adjacency matrices are closely related to our method. Franceschi et al. \cite{DBLP:journals/corr/abs-1903-11960} use bi-level program to first sample adjacency matrix and then learn the parameters for the graph by minimizing inner and outer objectives. In contrast, in the direct learning way, we parameterize the adjacency matrix as a learnable matrix and jointly learn the adjacency matrix and graph parameters via gradient descent. Also, we provide more ways to define adjacency matrix. As for graph pooling, the DiffPool \cite{diffpool} learns a differentiable soft cluster strategy for nodes using node embedding, mapping nodes to a set of clusters.  In contrast, we utilize the adjacency matrix to learn the cluster assignment matrix for nodes, which considers the link similarity of nodes.  Compared to using node embedding to learn cluster assignment matrix, using adjacency matrix is more intuitive and simple.
StructPool \cite{Yuan2020StructPoolSG} uses conditional random fields to capture the high-order structural relationships among different nodes to learn a node cluster assignment matrix based on the node features, where the adjacency matrix is used to find the topological information of the graph. In contrast, we directly utilize the link information in adjacency matrix to learn the cluster assignment matrix.

\begin{figure*}
\setlength{\belowcaptionskip}{-0.3cm}
\centering
\includegraphics[scale=0.25]{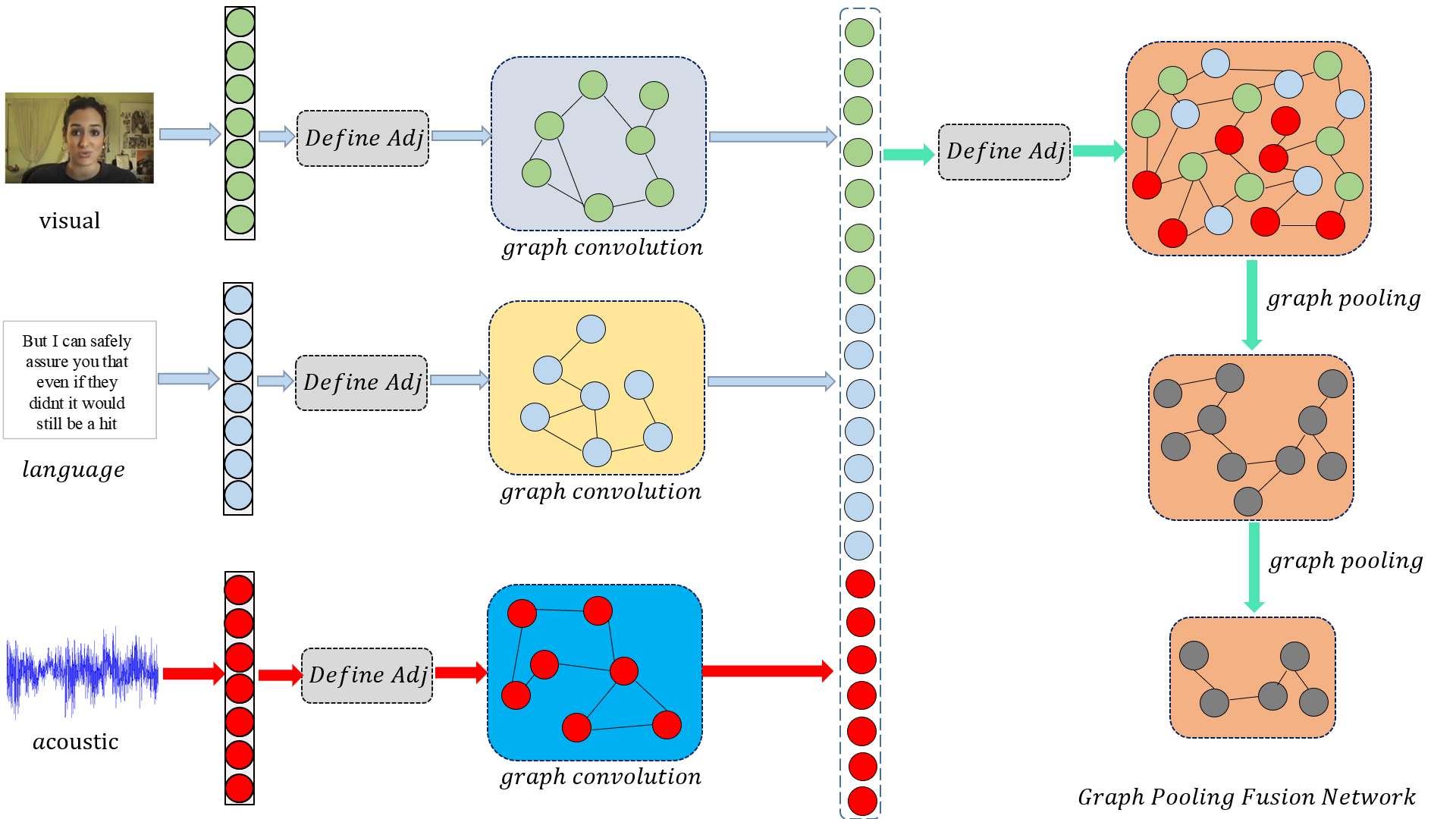}
\caption{\label{1}\textbf{The Schematic Diagram of Multimodal Graph.}
}
\end{figure*}

\section{\textbf{Model Architecture}}\label{sec:Algorithm}
In this section, we describe Multimodal Graph in detail, with its diagram illustrated in Fig.~\ref{1}. Multimodal Graph is hierarchically structured to cater to two stages, i.e., intra-modal and inter-modal dynamics learning. In the first stage, we propose to employ a graph convolutional network (GCN) for each modality. In the second, a graph pooling fusion network (GPFN) is devised for capturing cross-modal interactions. As an early attempt to employ GCN for sequential learning, the architecture of Multimodal Graph is free of recurrent neural networks (RNN) which are commonly used for sequential learning but subject to a number of limitations such as slowing inferring speed. Extensive discussion on applying this graph-oriented approach to sequential data is also provided in this section, including the definition of adjacency matrices.

\subsection{\textbf{Notations and Task Definition}}

Our downstream task is multimodal sentiment analysis. The input to the model is an utterance \cite{Olson1977From}, which is a segment of a video bounded by pauses and breaths. Each utterance has three modalities, i.e., acoustic ($a$), visual ($v$), and language ($l$) modalities. The sequences of acoustic, visual, and language modalities are denoted as $\bm{N_a} \in \mathbb{R}^{T_a \times d_a}$, $\bm{N_v} \in \mathbb{R}^{T_v \times d_v}$, and $\bm{N_l} \in \mathbb{R}^{T_l \times d_l}$, respectively. We aim to predict the sentiment score using the unimodal sequences.

To follow the notational convention in graph neural networks, a graph can be denoted as $G = (N, E)$, where $N= \{n_1, n_2, ..., n_T \}$ is a set of nodes, and $E\subseteq N\times N$ refers to the set of observed edges. The set of node embeddings is denoted as $\bm{N}\in \mathbb{R}^{T \times d}$ where $d$ refers to the dimensionality of each node embedding and $T$ is the number of nodes. The edges can also be described using the adjacency matrix $\bm{A} \in \mathbb{R}^{T \times T}$ where each element of $\bm{A}$ is non-negative and $A_{i, j}=0$ means that nodes $i$ and $j$ are not directly connected.

Graph convolutional network (GCN) is basically a convolutional operation on the nodes that are directly connected. The $k^{th}$ iteration of the general GCN can be described using the following equations:
\begin{equation}
  \bm{a}_{i}^{k-1}=\text{AGGREGATE}({\bm{N}_j^{k-1}; j\in \eta (i)})
\end{equation}
\begin{equation}
 \bm{N}_i^{k} = \text{COMBINE}(\bm{N}_i^{k-1},  \bm{a}_{i}^{k-1})
\end{equation}
where $ \eta (i) $ represents the set of 1-hop neighbors of node $i$, the $\text{AGGREGATE}$ function aggregates information from the 1-hop neighbors of node $i$ and output the aggregation representation $\bm{a}_{i}^{k-1}$, and the $\text{COMBINE}$ function combines the information of node $i$ and its aggregation information to update the representation of node $i$.

\subsection{\textbf{Unimodal Graph Convolution}}\label{sec:MSL}
To address the first challenge, i.e., learn intra-modal interactions, we utilize three unimodal GCNs on the unimodal sequences, each for one modality.
Intra-modal interactions are essential in multimodal language analysis which involves the modality-specific sentiment information. The majority of prior works tend to leverage RNNs to explore intra-modal dynamics \cite{RMFN,Zadeh2018Memory,mosei,RAVEN,MFRM}, which are prone to some limitations as stated in the Introduction section. In contrast, we propose to use GCNs to learn unimodal high-level representations for each modality, which allows parallel computation in the time dimension and is capable of learning long-term temporal contextual information with few layers.

We define the acoustic, visual, and language graphs as $G^a = (N^a, E^a)$, $G^v = (N^v, E^v)$, and  $G^l = (N^l, E^l)$, respectively. Taking the acoustic graph as an example, $N^a= \{n_1^a, n_2^a, ..., n_{T_a}^a \}$ is a set of acoustic nodes, $E^a\subseteq N^a\times N^a$ refers to the set of edges that directly connects the acoustic nodes. The acoustic node embedding is denoted as $\bm{N_a}\in \mathbb{R}^{T_a \times d_a}$, which is the input acoustic sequence in our setting.

However, unlike graph-structured data, a unimodal sequence does not have an adjacency matrix that determines the graph topology. Hence, one major problem is to define the adjacency matrix in the unimodal sequence such that it effectively reflects the connection between nodes (time steps).
Intuitively, two nodes are assumed to be connected if they are close in terms of the feature embedding. Therefore, we can measure the similarity between the embeddings of each two nodes to determine whether they are neighbors. Here we use a simple cross-node attention mechanism to determine the correlation between nodes and thus define the adjacency matrix for unimodal sequences. The equations are shown below:
\begin{equation}
\setlength{\abovedisplayskip}{3pt}
\setlength{\belowdisplayskip}{3pt}
 \bm{\hat{A}}=f[f(\bm{W}_1 \bm{Q}) f((\bm{W}_2 \bm{P})^R)]
\end{equation}
\begin{equation}
\setlength{\abovedisplayskip}{3pt}
\setlength{\belowdisplayskip}{3pt}
 A_{i,j}=\frac {\hat{A}_{i,j}} {\sum_{v \in N}\hat{A}_{i,v}+\epsilon}
\end{equation}
where $ \bm{Q}\!=\!f(\bm{W}_q\bm{N}) \in \mathbb{R}^{T \times d}, \bm{P}\!=\!f(\bm{W}_p\bm{N}) \in \mathbb{R}^{T \times d} $, and $\bm{W}_1, \bm{W}_2, \bm{W}_q, \bm{W}_p \in \mathbb{R}^{d \times d}$ are learnable matrices. We omit the superscript that indicates modality in the symbols and equations for conciseness. $f$ is the nonlinear activation function to increase the nonlinear expressive power of the model and $R$ denotes the matrix transpose operation. Note that we apply $ReLU$ as our activation function in all the equations such that the negative links between nodes can be effectively filtered out. The equations are the same for each single modality except that the node embedding to learn the adjacency matrix is different.
This approach to constructing an adjacency matrix for a temporal sequence is learnable with parameters, and meanwhile it is instance-specific as it considers the various relatedness among nodes (time steps), as opposed to directly setting all the matrix elements as learnable parameters (we will discuss it in section~\ref{sec:adj}). Hence, we term this approach indirect learning. We claim that this instance-specific and learnable approach can capture more relatedness information on the nodes can generate more favourable performance, as shown in Section \ref{sec:adj_matrix}.

From this definition of adjacency matrix for sequential data, it can be seen that even if the two nodes are distant apart in the time dimension, they can still be directly connected if they are considered related.
Therefore, compared to RNN variants, GCN can effectively learn long-term temporal dependency with fewer layers, which makes it a suitable alternative to learn long temporal sequences.
Although the indirect learning method has some similarity to Transformer \cite{transformer} in avoiding recurrence and learning long-term dependency, this approach is different from it in several aspects.
Firstly, Transformer uses $softmax$ as activation function, which means that each time step is associated with all time steps.
In contrast, our method is better-targeted in that it is able to filter out the time steps that have no direct connection, and automatically detect the one-hop neighbors for each node.
Moreover, after finding the neighbors for each time step, we use a GCN such as GIN \cite{DIN} to aggregate the information of the neighbors, and this operation could be quiet different from the feed forward network of Transformer.

Note that in common graph definition, the elements in the adjacency matrix are often binary and restricted to either 0 or 1, which denotes no/one direct connection, respectively. However, we dispense with this restriction and formulate the elements to be continuous, with a larger value indicating a closer relation between two nodes, and vice versa. This can be interpreted as multiplying an attention mask matrix to the conventional adjacency matrix. We justify the use of such continuous soft weights in adjacency matrix in Section~\ref{sec:weight}. Apart from the indirect learning method, we additionally provide multiple ways to construct the adjacency matrix, which will be discussed in Section~\ref{sec:adj}.

After obtaining the adjacency matrix, the definition of $\text{COMBINE}$ and $\text{AGGREGATE}$ functions in GCN could have many choices.
It is worth mentioning that the unimodal graphs are independent of the concrete GCN model. In other words, we can integrate any GCN model into our unimodal graph. In practice, we compare the performance of Graph Isomorphism Network (GIN) \cite{DIN}, Graph Attention Network (GAT) \cite{GAT}, GraphSAGE \cite{GraphSAGE} and DiffPool \cite{diffpool} in our experiment. Specifically, we use GraphSAGE with mean pooling \cite{GraphSAGE} as the default GCN in this paper, and the equations for the $k^{th}$ iteration of GraphSAGE is shown below:
\begin{equation}
\label{eq1111}
 \bm{n}^k_i=f(  \bm{D}^{-1} (\bm{A}+\bm{I})  \bm{N}^{k-1} \bm{W}^k)_i
\end{equation}

\begin{equation}
\label{eq6}
 \bm{N}^k_i= \frac{\bm{n}^k_i}{||\bm{n}^k_i||_2}
\end{equation}
where $f$ is the non-linear activation function for which we use $ReLU$ in our experiment, $\bm{N}^{k-1}_i$ is the hidden representation for node $i$ at iteration $k-1$, and $\bm{W}^k \in \mathbb{R}^{d \times d}$ is the parameter matrix. $\bm{D}$ is the  diagonal degree matrix of $\bm{A}$ where $D_{ii} = \sum_j A_{ij} $ and $D_{ij}=0 \ (i\neq j)$. The diagonal degree matrix $\bm{D}$ is added to perform mean pooling, and the identity matrix $\bm{I}$ is added to the adjacency matrix $\bm{A}$ to perform self-loop operation. Normalization is done in Eq.~\ref{eq6}. We use the same convolution structure for each modality, and the superscript that indicates modality is omitted for conciseness. To obtain the final unimodal representations, the hidden representations $\bm{N}^k_i$ for each iteration is concatenated and sent to the fully connected layers:
\begin{equation}
 \bm{N}'_i \leftarrow \oplus\bm{N}^k_i,\ k \in \{1,2,..,l\}
\end{equation}
\begin{equation}
 \bm{N}_i = f(\bm{W}_o \bm{N}'_i + \bm{b}_o)
\end{equation}
where $\bm{N}_i$ is the final representation of node $i$, $\oplus$ denotes concatenation, $l$ is the number of iterations, and $\bm{W}_o$ and $ \bm{b}_o$ are the weight matrix and bias for the fully connected layers, respectively.

With graph convolution, the three GCNs for three respective unimodal sequences can explore the intra-modal interactions effectively. Unlike the commonly used RNN variants which are subject to a number of issues such as gradient vanishing/explosion, forgetting problem and slow inferring speed, GCN abandons the recurrence and can operate in parallel, which is more efficient in terms of time complexity and can learn longer temporal dependency. More importantly, it detects the immediate (one-hop) neighbors for each time step and computes their relatedness, which is better-targeted compared to Transformer. Extensive experiments are conducted in Section \ref{compare_RNN} to show the superior performance of GCN in modeling sequential data.

\subsection{\textbf{Cross-modal Graph Pooling Fusion Network}}\label{sec:LIL}
After exploring intra-modal interactions for each modality, the second challenge is to model inter-modal dynamics and fuse the cross-modal nodes.  Considering that we focus on unaligned multimodal sequences, the common word-level fusion cannot be achieved. This means that our fusion network should learn the interactions among various nodes (time steps) from multiple modalities, rather than fuse the features from three modalities at each specific time step.
To this end, we devise a graph pooling fusion network (GPFN) over the unimodal graphs. GPFN learns to aggregate the multimodal nodes to learn high-level graph representations hierarchically. In GPFN, we introduce max/mean graph pooling and analyze their rationality. Additionally, different from other graph pooling methods that mainly utilize the node embedding to learn a cluster assignment matrix \cite{diffpool,Yuan2020StructPoolSG}, we propose link similarity pooling to learn a cluster assignment matrix using the link information of adjacency matrix, which is more intuitive and interpretable.

\begin{figure}
\setlength{\belowcaptionskip}{-0.3cm}
\centering
\includegraphics[scale=0.28]{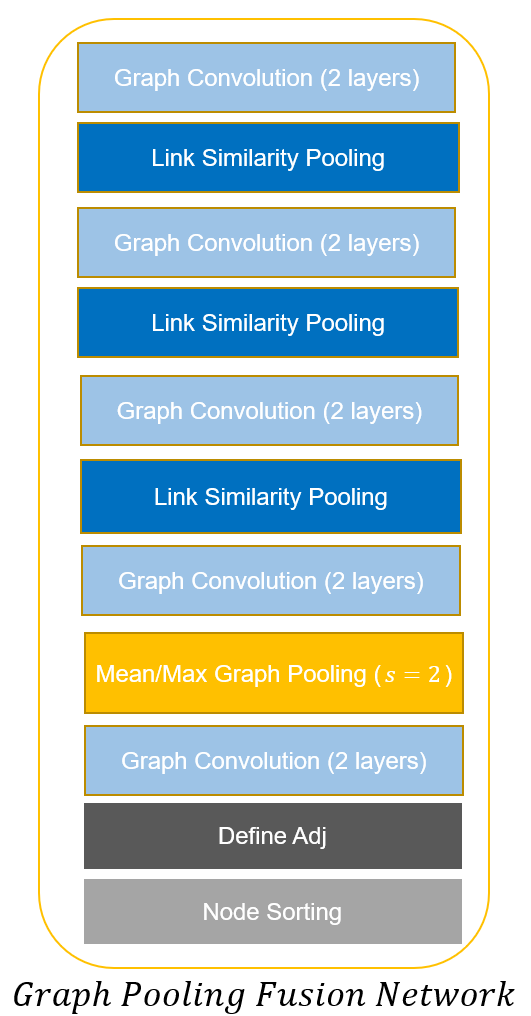}
\caption{\label{1111}\textbf{The Detailed Structure of GPFN. $s$ denotes pooling size of max/mean pooling.}
}
\end{figure}

The overall architecture of GPFN is illustrated in Fig.~\ref{1111}. As shown in the figure, our GPFN comprises of node sorting, adjacency matrix definition, graph convolution, mean/max graph pooling, and link similarity pooling. To retain consistence, the graph convolution framework and the adjacency matrix definition method are the same as those in unimodal graphs. In the followling subsections, we will introduce node sorting, max/mean graph pooling, and link similarity pooling, respectively.

\subsubsection{\textbf{Node Sorting}}

Firstly, we need to arrange the nodes to determine the order of the nodes from the three unimodal graphs and obtain the multimodal sequence. An intuition here is to sort nodes from three modalities according to time dimension such that the nodes from neighboring time steps are closer to each other. However, this requires that the time dimensions of different modalities are explicitly aligned. Hence, we simply concatenate the nodes for different modalities, and let the model automatically learn to aggregate these nodes. The node set $N^m$ can be described as $N^m= \{n_1^l, n_2^l,..., n_{T_l}^l,  n_1^a, n_2^a, ..., n_{T_a}^a,  n_1^v, n_2^v, ..., n_{T_v}^v \}=\{n_1^m, n_2^m,..., n_{T_l+T_a+T_v}^m \}$.

Since features from different modalities are highly heterogeneous, the interactions between the cross-modal nodes are much more complex than those of the nodes from a single modality. Therefore, we learn the interactions between the unimodal sequences via graph pooling. In this way, the model can automatically learn the complex interactions between the heterogeneous modalities by associating the related nodes between these three modalities.

\subsubsection{\textbf{Mean/Max Graph Pooling}}

The initial adjacency matrix is computed in the same way as the unimodal graphs. With graph pooling, the related nodes are automatically fused so that the interactions can be explored and the new adjacency matrix can be obtained. In the GPFN, we provide two pooling approaches to aggregate the nodes. The first kind is the simple max pooling and mean pooling. For the adjacency matrix, 2D mean/max pooling is applied, while for the node embeddings, 1D mean/max pooling is applied.
The equations for mean pooling are given as follows:
\begin{equation}
  \bm{N}_{i}^{k}=\frac{\sum_{g=0}^{s-1} \bm{N}_{s\cdot i - g}^{k-1}}{s}
\end{equation}
\begin{equation}
\label{eq7}
  A_{i,j}^{k}=\frac{\sum_{g=0}^{s-1}\sum_{m=0}^{s-1} A_{s\cdot i - g, s\cdot j - m}^{k-1}}{s^2}
\end{equation}
and for max pooling:
\begin{equation}
\label{eq1111}
  \bm{N}_{i}^{k}=\text{max}(\{\bm{N}_{s\cdot i - g}^{k-1}\ |\ 0\leq g\leq s-1\})
\end{equation}
\begin{equation}
  A_{i,j}^{k}=\text{max}(\{A_{s\cdot i - g, s\cdot j - m}^{k-1}\ |\ 0\leq g\leq s-1, 0\leq m\leq s-1\})
\end{equation}
where $s$ is the pooling size, $\bm{N}_{i}^{k}$ is the updated node embedding for node $i$ at iteration $k$, and $\cdot$ denotes scalar multiplication. Note that in Eq~\ref{eq1111}, the max pooling operation is element-wise.

The mean/max pooling is meaningful because the nodes in multimodal sequence are concatenated according to the time dimension of each unimodal sequence, and thus the neighboring nodes are closely related in time dimension and considered to be fusible  (but we need to carefully determine the pooling size $s$ to avoid fusing the nodes from different modalities). Moreover, we have the following observation:

\textbf{Observation 1}: If nodes $x$ and $y$ are 1-hop neighbors (directly connected), then they are 1-hop or 0-hop neighbors after mean/max graph pooling, where 0-hop neighbors mean that node $x$ and $y$ are merged into the same node after graph pooling.

\textbf{Proof}:  We do not consider the case where they are merged into the same node after graph pooling since it is obvious by definition. Since the nodes $x$ and $y$ are the 1-hop neighbors at iteration $k-1$, we have $A_{x,y}^{k-1}>0$ or $A_{y,x}^{k-1}>0$ (the adjacency matrix is not always symmetric). Without loss of generalization, we assume $A_{x,y}^{k-1}>0$. Let us assume that $x$ and $y$ are merged into nodes $x'$ and $y'$, respectively, according to the definition of 1D pooling, we have $x'=ceil(x/s)$, $y'=ceil(y/s)$ (where $ceil$ means rounding up to an integer). According to Eq.~\ref{eq7}, we have
$A_{x',y'}^{k}=\frac{\sum_{g=0}^{s-1}\sum_{m=0}^{s-1} A_{s\cdot x' - g, s\cdot y' - m}^{k-1}}{s^2}$  (take mean pooling as example). Obviously, from $x'=ceil(x/s)$, we have $x' -1 < x/s \leq x' $, which implies that $s\cdot x' - s < x \leq s\cdot x' $. Given that $x', s, x$ are all integers, we have $x \in [s\cdot x' - s+1, s\cdot x'] $. Since $0\leq g\leq s-1$, we further have  $ s\cdot x'-s + 1\leq s\cdot x' - g \leq s\cdot x'$, which implies that $x \in \{s\cdot x' - g\ |\ 0\leq g\leq s-1 \}$. Similarly, we have $y \in \{s\cdot y' - m\ |\ 0\leq m\leq s-1 \}$. This indicates that at least one of the elements in $A_{s\cdot x' - g, s\cdot y' - m}^{k-1}$ is greater than zero. Therefore we have $A_{x',y'}^{k}>0$, which implies that nodes $x'$ and $y'$ are 1-hop neighbors.

The above property of mean/max pooling suggests that they are reasonable approaches for graph pooling, and indicates a principle for us to manually design graph pooling algorithms: once neighbors, always neighbors.

Although the mean/max graph pooling may be effective in fusing nodes from the same modality given that we can utilize the time information in the unimodal sequence, it is not a desirable method for fusing nodes from different modalities in unaligned multimodal sequences. Therefore, we need a learnable method that can effectively cluster the heterogeneous cross-modal nodes. To this end, we propose link similarity pooling in the following section.

\subsubsection{\textbf{Link Similarity Pooling}}
Apart from the mean/max graph pooling, we devise a learnable graph pooling method, named link similarity pooling, to leverage the link information (topology information) in the adjacency matrix to learn the node cluster assignment matrix. Different from DiffPool \cite{diffpool} that uses the node embedding to learn a cluster assignment matrix, using adjacency matrix is more intuitive and interpretable.

We define the link similarity pooling in the following equations:
\begin{equation}
  \bm{Z}' = \bm{A}\bm{A}^R
\end{equation}
\begin{equation}
  \bm{Z} = f[(\bm{Z}' + \bm{A})\bm{W}_z]
\end{equation}
where $R$ denotes matrix transposition, $\bm{W}_z \in \mathbb{R}^{T \times T'}$ is the transfer parametric matrix, and $\bm{Z}\in \mathbb{R}^{T \times T'}$ is the final node cluster assignment matrix that maps the $T$ nodes into $T'$ nodes.
The first equation measures the similarity score of two nodes by calculating the inner product of their respective association intensity to their common neighbors. In this way, a greater extent to which a pair of nodes have similar distribution of association intensity on their shared neighbors leads to a larger score. It is also noteworthy that this score is weighted because the elements in the adjacency matrix are soft (continuous) and not restricted to be binary.
For example, if nodes $x$ and $y$ have more common neighbors and their linked values with the common neighbor are larger, then $Z'_{x,y}$ is larger. In Fact, we have $Z'_{x,y}=\sum_{c\in CN}A_{x,c}\cdot A_{y,c}$ where $CN$ denotes the set of common neighbors between $x$ and $y$. The second equation adds the original adjacency matrix to the link similarity information, which means that $Z_{x,y}$ is larger if $x$ and $y$ are 1-hop neighbors. This is reasonable because if two nodes are neighbors, then they are considered to be similar and thereby can be fused with a high possibility.

After obtaining the node cluster assignment matrix $\bm{Z}$, we use it to learn the updated adjacency matrix and node embedding. The equations are presented as follows:
\begin{equation}
  \bm{S} = f(\bm{N}\bm{W}_s)
\end{equation}
\begin{equation}
  \bm{N}_{\text{update}} = \bm{Z}^R\bm{S}, \ \ \bm{A}_{\text{update}} = \bm{Z}^R\bm{A}\bm{Z}
\end{equation}
where $\bm{W}_s \in \mathbb{R}^{d \times d'}$ is a learnable parameter matrix, $\bm{N}_{\text{update}}\in \mathbb{R}^{T' \times d'}$ and $\bm{A}_{\text{update}}\in \mathbb{R}^{T' \times T'}$ denote the updated node embedding and adjacency matrix, respectively.

The reason why we do not use the node embedding is that the adjacency matrix is originally derived from the node embedding, and therefore adjacency matrix contains part of the information in node embedding as well as the topology information. Therefore, we assume that using the adjacency matrix is sufficient to learn the node cluster matrix.

After GPFN, we average the node embeddings as the representation of GPFN, and we concatenate the representation of GPFN with the averaged node embedding from the three unimodal graphs respectively to obtain the final graph representation. Several fully connected layers are applied on the final graph representation to infer the final sentiment decision.

\section{\textbf{Discussion on the Adjacency Matrix}}\label{sec:adj}
Previously we define an indirect learning method to construct the adjacency matrix for sequential data, which is instance-specific and learnable. In this section, we aim to explore other reasonable methods to construct the adjacency matrix. Specifically, we present three other ways to define the adjacency matrix, namely Generalized Diagonal Matrix, KNN-based adjacency matrix, and a direct learning method.

\subsection{\textbf{Generalized Diagonal Matrix}}
Intuitively, the neighboring time slices are more related to each other. Therefore, we define a generalized diagonal matrix (GDM) to reflect this point:
\begin{equation}       
\setlength{\abovedisplayskip}{3pt}
\setlength{\belowdisplayskip}{3pt}
\left(                 
  \begin{array}{cccccccc}   
    1 & \lambda & \lambda^{2} & ... & \lambda^{n-1} & 0 & ... & 0\\
    \lambda & 1 & \lambda & ... & \lambda^{n-2} & \lambda^{n-1} & ... & 0\\
    \vdots & \ddots & \ddots & \ddots & \ddots & \ddots & \ddots & \vdots\\
     0 & ... & 0 & \lambda^{n-1} & ... & \lambda & 1 & \lambda\\
    0 & ... & 0 & 0 & ... & \lambda^{2} & \lambda & 1\\
  \end{array}
\right)                 
\end{equation}
where $\lambda$ is the attenuation factor which is set to 2 in our experiment, and $n$ is the truncation factor.
The value on the diagonal of the GDM is 1, and each element is decayed by a factor of $\frac{1}{\lambda}$ centered on the diagonal, which means the correlation between them is reduced with the increase of distance.
When the distance to the diagonal element is larger than $n$, the value becomes zero, which means that the distant nodes no longer have direct (1-hop) connection. Nevertheless, we can stack many layers so that each node can still have the overall view of the input sequence. Obviously, for a sequence of length $T$, we need to stack $ceil((T-1)/(n-1))$ layers such that each node can incorporate information from all the nodes, where $ceil$ means rounding up to an integer.

A  huge advantage of GDM lies in that it eschews the complex computation for finding an adjacency matrix. Actually, GDM functions like the kernels in TCN \cite{TCN,Trellis,TCNN}, but unlike TCN kernels, it is predefined instead of being obtained through learning (we leave the learning part to the parameters of GCN).
The main differences between GCN and TCN will be discussed in section~\ref{sec:Analysis}.
In addition, we also implement a fully connected adjacency matrix whose values are all ones to make a comparison. Compared to the indirect learning method, this GDM method is intuitive and fits the empirical pattern of sequence modeling, but it is neither instance-specific nor learnable. GDM serves as a reasonable baseline in the exploration of adjacency matrix for sequential data.


\subsection{\textbf{KNN-based method}}

Another simple but effective non-parametric approach to finding an adjacency matrix in a sequence is to apply K-nearest neighbor (KNN) algorithm to define the 1-hop neighbors of each node, which has also been evaluated in \cite{DBLP:journals/corr/abs-1903-11960}. KNN is based on Euclidean distance, and for each node, we select the nodes with short distance as the 1-hop neighbors of this node. The equations can be described as below:

\begin{equation}
d_{ij}=\frac {1}{\text{Eur}({\bm{N}_j; \bm{N}_i})+\epsilon}
\end{equation}

\begin{equation}
\label{e18}
\hat{A}_{i,j}= \text{ReLU}(d_{ij}- \alpha \times \frac{\sum_j d_{ij}}{T})
\end{equation}

\begin{equation}
\label{e19}
\setlength{\abovedisplayskip}{0pt}
\setlength{\belowdisplayskip}{0pt}
 A_{i,j}=\frac {\hat{A}_{i,j}} {\sum_{v \in N}\hat{A}_{i,v}}
\end{equation}
where $\text{Eur}$ denotes the Euclidean distance, $d_{ij}$ denotes the `similarity' of  node $j$ to node $i$, $\epsilon$ denotes a positive scalar to prevent division by zero, and $\alpha$ is a scalar that controls to what extent the weak links can be filtered out. Eq.~\ref{e18} filters out the links that have no strong connection and Eq.~\ref{e19} denotes simple normalization. Note that by applying Eq.~\ref{e18}, we do not select an exact number of $k$ nearest neighbors for each node, but we allow a variable number of neighbors for each node, as long as the links between the node and its neighbors meet the threshold. Therefore, our KNN-based method is not a strict $k$-nearest neighbor algorithm.
A disadvantage of KNN-based method is that it has to compute the Euclidean distance of each two nodes for each instance, making it more time-consuming in training and inferring periods compared to GDM. By definition, this KNN-based method is instance-specific, but it is not a learnable one, serving as one reasonable comparative method to the indirect learning method.

\subsection{\textbf{Direct Learning}}
Another intuitive way to construct an adjacency matrix is to learn the matrix directly in the training process.
In this method, the adjacency matrix is parameterized as a learnable matrix $\hat{\bm{A}}\in \mathbb{R}^{T \times T}$, and a ReLU function is applied to activate the learned matrix: $\bm{A}=\text{ReLU}(\hat{\bm{A}})$ such that the linked values between nodes are non-negative.

To obtain a more representative adjacency matrix, we add a regularization term to the learned adjacency matrix, as shown below:
\begin{equation}
\setlength{\abovedisplayskip}{3pt}
\setlength{\belowdisplayskip}{3pt}
 \ell\!=\!\!\sum_{i\in N}(\sum_{j \in N}\hat{A}_{i,j})^2 +\! \sum_{i\in N}(\sum_{j \in N}\text{ReLU}(\hat{A}_{i,j})\!-\!\gamma)^2
\end{equation}
where the first term forces the sum of linked values of each node to equal zero, which can prevent the learned matrix from becoming a fully-connected matrix; the second term restricts the sum of the positive linked values of each node to approximate a given positive scalar $\gamma$ so as to prevent it from degenerating into an all-zero matrix. $\ell$ is optimized via gradient descent. Note that the learnt adjacency matrix is shared across instances (instance-independent), which may not be expressive enough compared to the instance-dependent learning method but is computationally efficient.

\section{\textbf{Links to TCN}}\label{sec:Analysis}
TCN is a rising star in natural language processing (NLP) tasks that outperforms RNN variants on a variety of tasks \cite{TCN,Trellis,TCNN,pmlr-v70-gehring17a}. TCN can also learn longer temporal dependency and operate in parallel. In this section, we present a brief discussion on the main difference of GCN and TCN.
The basic functions of TCN and GCN can be expressed as:
\begin{equation}
\setlength{\abovedisplayskip}{3pt}
\setlength{\belowdisplayskip}{3pt}
  \bm{x}^{tcn}_t=\bm{f}\star\bm{N}_t=\sum_{i=0}^{k-1} \bm{f}_i \bm{N}_{t-i},\ 1\leq t \leq T, t-i>0
\end{equation}
\begin{equation}
\label{eq10}
\setlength{\abovedisplayskip}{3pt}
\setlength{\belowdisplayskip}{3pt}
  \bm{X}^{gcn}=\bm{A}\bm{N}\bm{W_{gcn}},\ \bm{x}^{gcn}_t=\bm{A_{r_t}}\bm{N}\bm{W_{gcn}}
\end{equation}
where $\bm{f}\in \mathbb{R}^{k \times d_{in} \times d_{out}}$ is the convolutional filter ($\bm{f}_i\in \mathbb{R}^{d_{in} \times d_{out}}$ is the $i^{th}$ dimension of the convolutional kernel, and $k$ is the kernel size), $\bm{A}\in \mathbb{R}^{T \times T}$ is the adjacency matrix ($\bm{A_{r_t}}$ denotes the $t^{th}$ row of $\bm{A}$), $\bm{N}\in \mathbb{R}^{T \times d_{in}}$ is the sequence embedding, and $\bm{W_{gcn}}\in \mathbb{R}^{d_{in} \times d_{out}}$ is the parameter for GCN. As can be seen, a main difference of TCN and GCN is that the filter $\bm{f}$ in TCN is shared across all time steps of the sequence (independent of $t$), while each row of $\bm{A}$ in GCN corresponds to one node (time step) and it often varies with different nodes. In other words, $\bm{A}$ is more interpretable and specific for each time step, and thus is more representative. Moreover, the convolutional filter $\bm{f}$ is learned automatically via gradient descent, while $\bm{A}$ can be pre-defined empirically. Furthermore, in GCN, we can connect two distant nodes (time steps) via $\bm{A}$ if the corresponding element in $\bm{A}$ is positive. In this way, longer-term dependency can be effectively learnt with few layers.
In contrast, although the dilated temporal convolution  \cite{TCN} can also link two distant time steps with low layers via dilated convolution, it is inflexible and cannot automatically learn the long-term connection, because the dilated rate in the dilated temporal convolution has to be manually predefined.

\begin{figure*}
\setlength{\belowcaptionskip}{-0.3cm}
\centering
\includegraphics[scale=0.25]{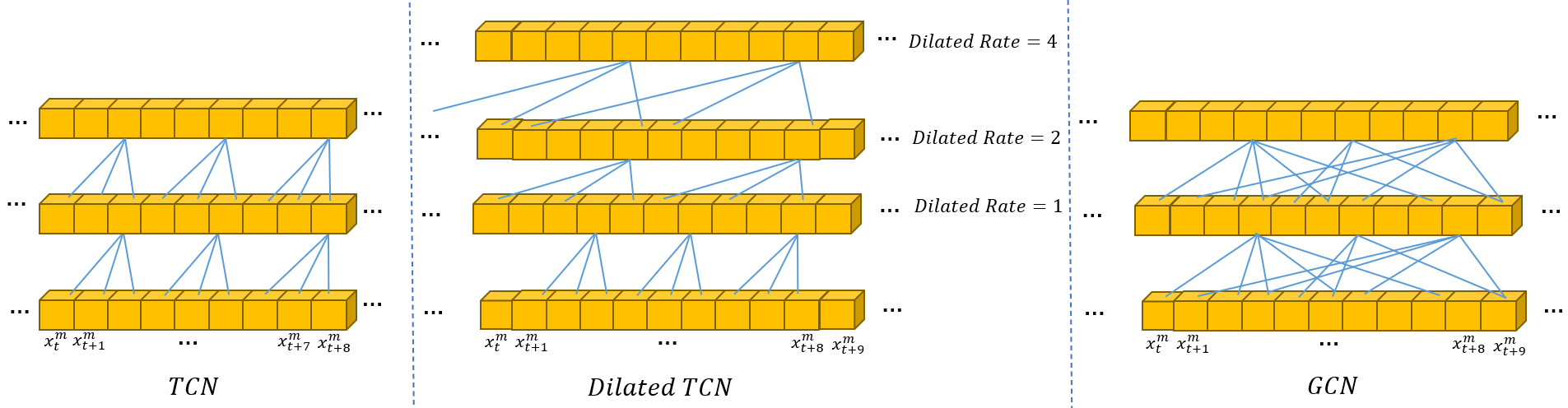}
\caption{\label{111}\textbf{The Schematic Diagram of TCN, Dilated TCN and GCN.}
}
\end{figure*}


The schematic diagrams of TCN, dilated TCN and GCN are provided in Fig.~\ref{111}. As shown in the figure, TCN and dilated TCN are subject to some fixed convolutional patterns which are rather inflexible. Instead, the graph convolution can be very flexible and expressive which associates the related time steps automatically.

\section{\textbf{Justification of the Use of Soft Weights in Adjacency Matrix}}\label{sec:weight}
In our definition of adjacency matrix, the values are not binary and restricted to 0 and 1 as in vanilla GCN.
One might raise the question that if binary adjacency matrices are used, can the weights of the GCN layers compensate for the binary nature of the adjacency matrix and scale them internally? In other words, is it really necessary to use soft weights in the adjacency matrix? In this section, we mathematically describe how adjacency matrices and weights in GCN interact to clarify the advantage of using soft weights.

According to Eq.~\ref{eq10}, we have (we omit the $gcn$ superscript and subscript for conciseness):
\begin{equation}
\begin{split}
\bm{X}&=\bm{ANW}=(\bm{AN})\bm{W}=\bm{N'W}\\
      &=[\bm{N'_1};\ \bm{N'_2};\ ...;\ \bm{N'_T}]\ \bm{W}\\
      &=[\bm{N'_1W};\ \bm{N'_2W};\ ...;\ \bm{N'_TW}]
\end{split}
\end{equation}
where $\bm{N'}=\bm{AN}$,  $\bm{N'_i} = \sum_{k=1}^T (A_{ik} \bm{N_k})\in \mathbb{R}^{d_{in}}$ is a row vector. Then we have:
\begin{equation}
\begin{split}
\bm{X_i} &= \bm{N'_i W} = [\sum_{k=1}^T (A_{ik} \bm{N_k})]\bm{W} \\
         & = \sum_{k=1}^T [A_{ik} (\bm{N_k W})]
\end{split}
\end{equation}
where $\bm{X_i}$ is the updated embedding for node $i$.
At this point, we can see that $\bm{N_k W}$ is shared across all $\bm{X_i}$, and $A_{ik}$ is unique for each $\bm{X_i}$. If the values in $\bm{W}$ change, it will have a global impact on all the node embeddings, while the value of $A_{ik}$ only affects node $i$. Therefore, $\bm{W}$ cannot compensate for $A_{ik}$, and they have different functions. Furthermore, $A_{ik}$ can be regraded as the weight for $\bm{W}$ if $\bm{N}$ is considered to be fixed. Without loss of generalization, assuming that $\bm{N_k}$ is one-hot encoded where $N_{kk} = 1$ and $N_{kg} = 0$ ($g \neq k$), then we have $\bm{X_i} = \sum_{k=1}^T  [A_{ik} \bm{W_k}]$, which further suggests that $\bm{A}$ and $\bm{W}$ function in different ways and they cannot compensate for each other. If $A_{ik}\in \{0,1\}$, then the updated node embedding will be the simple combination of $\bm{N_k W}$, which we think is not expressive enough to distinguish different node embeddings. Moreover, the binary adjacency matrices treat all the neighbors equally and cannot identify which two nodes are closer to each other. Therefore, using soft weights in adjacency matrix is more expressive.

\section{\textbf{Experiments}}\label{sec:Experiments}

Multimodal Graph is evaluated on two popular datasets for multimodal sentiment analysis. In this section, we focus on the following questions: 1) Does GCN perform favorably to TCN and RNN variants? 2) Does Multimodal Graph achieve state-of-the-art performance on multimodal sentiment analysis? 3) What kind of adjacent matrix performs best? 4) What are the attributes of the learned adjacent matrix? 5) What kind of graph neural network performs best in our Multimodal Graph?

\subsection{Datasets}

\subsubsection{\textbf{CMU-MOSI}}
CMU-MOSI \cite{Zadeh2016Multimodal} is a widely-used dataset for multimodal sentiment analysis. It contains 93 videos in total, and each video is divided into 62 utterances at most. The intensity of emotion ranges within [-3,3], where -3 indicates the strongest negative emotion, and +3 the strongest positive. We evaluate model's performance using various metrics, in agreement with those employed in prior works such as RAVEN \cite{RAVEN}. The metrics include 7-class accuracy (i.e., Acc7: sentiment score classification), binary accuracy (i.e., Acc2: positive or negative sentiments), F1 score, mean absolute error (MAE) of the score, and the correlation of the model's prediction with humans. To be consistent with prior works, we use 1284 utterances as training set and 686 utterances as testing set.

\subsubsection{\textbf{CMU-MOSEI}}
CMU-MOSEI \cite{mosei} is the largest multimodal language analysis dataset that contains a total number of 2928 videos. The dataset has been segmented at the utterance level, and each utterance has been scored on two levels: sentiment ranging between [-3, 3], and emotion with six different values. In our experiment, the evaluated metrics for CMU-MOSEI are the same as those for CMU-MOSI dataset. We use 16265 utterances as training set and 4643 utterances as testing set.


\subsection{Baselines}
We compare the performance of Multimodal Graph with the following state-of-the-art multimodal machine learning models.

1) \textbf{Early Fusion LSTM} (\textbf{EF-LSTM}): EF-LSTM is the baseline fusion approach that utilizes the early fusion techniques to concatenate the input features of different modalities, and then send the concatenated multimodal features to an LSTM layer followed by fully connected layers. Since the EF-LSTM concatenates the features of all modalities at each time steps, it requires that the multimodal sequences are explicitly aligned and cannot handle unaligned multimodal sequence. As adopted in MulT \cite{mult}, connectionist temporal classification (CTC) \cite{CTC} is performed to process the unaligned sequences to obtain an approximately aligned sequences.  Specifically, EF-LSTM is trained to optimize the CTC alignment objective and the multimodal language analysis objective simultaneously.

2) \textbf{Late Fusion LSTM} (\textbf{LF-LSTM}): LF-LSTM is another baseline fusion method that uses an LSTM network for each modality to extract unimodal features and infer decision, and then combine the unimodal decisions by late fusion.

3) \textbf{Tensor Fusion Network} (\textbf{TFN}) \cite{Zadeh2017Tensor}: TFN adopts outer product to learn joint representation of three modalities, which can effectively preserve unimodal dynamics and explore bimodal and trimodal interactions. TFN is not a word-level alignment fusion method, and it can handle unaligned multimodal sequence for it summarizes the unimodal sequence into unimodal vector before fusion.



4) \textbf{Memory Fusion Network} (\textbf{MFN}) \cite{Zadeh2018Memory}: MFN uses systems of LSTM to learn intra-modal dynamics, and it implements delta-memory attention and multi-view gated memory network to fuse memories of LSTMs and explore inter-modal dynamics across time. Note that MFN is also a word-level fusion method, and thus we perform connectionist temporal classification (CTC) \cite{CTC} to process the unaligned sequences to obtain an approximately aligned sequences.



5) \textbf{Multimodal Transformer} (\textbf{MulT}) \cite{mult}: MulT applies 1 layer temporal convolutional networks to extract unimodal features, and then uses a Transformer \cite{transformer} to translate each two modalities. It can effectively model unaligned mutlimodal sequence. According to our experiments, it is currently the state-of-the-art model in multimodal language analysis.

\subsection{Experimental details}\label{sec:detail}
We develop our model on Pytorch with RTX2080Ti as GPU. We apply MAE as loss function with Adam \cite{Kingma2014Adam} of learning rate 0.001 as optimizer. For each baseline, following \cite{Gkoumas2021WhatMT}, we first fine-tune the model by performing fifty-times random grid search on the hyper-parameters and the hyper-parameter setting that reaches the best performance is saved. After the fine-tuning process, we train the model again with the best hyper-parameters for five times, and the final results are obtained by calculating the mean results.

For feature extraction, to make a fair comparison with baselines, we follow the setting of CMU-MultimodalSDK \footnote{https://github.com/A2Zadeh/CMU-MultimodalSDK}.  GloVe word embeddings \cite{pennington2014glove}, Facet \footnote{ iMotions 2017. https://imotions.com/},  and COVAREP \cite{Degottex2014COVAREP} are applied for extracting language, visual, and acoustic features respectively. The length of the language, acoustic, and visual sequence is 50, 500 and 500, respectively.

\subsection{Results and Discussions}

\subsubsection{\textbf{Comparison with TCN and RNN}}\label{compare_RNN}

\begin{table*}[t]
\centering
 \caption{ \label{t1}\textbf{ Comparison with RNN and TCN variants.} The GRU and LSTM models used here are bidirectional. Training Time means training time of the model per batch (the batch size and the environment are the same for all models).}
\resizebox{1.8\columnwidth}{!}{\begin{tabular}{c|c|c|c|c|c|c|c}
 \hline
    & Acc2 & Acc7  & F1 & MAE & Corr & Training Time (ms) & Parameters\\
 \hline
 GRU  & 80.4 & \textbf{50.5} & 80.5  & 0.611 & 0.669 & 933 & 1,018,811\\
LSTM  & 80.7 & 48.3 & 81.3  & 0.623 & 0.662 & 774 & \textbf{917,831}\\
 TCN  & 80.4 & 48.0  & 80.7 & 0.636 & 0.644 & \textbf{127} & 1,367,561\\
 \hline
 Multimodal Graph & \textbf{81.4} & 49.7 &  \textbf{81.7} & \textbf{0.608} & \textbf{0.675} & \textbf{125} & 1,225,400\\
 \hline
 \end{tabular}}
 \vspace{-0.3cm}
\end{table*}%

The mainstream approaches to processing sequences are RNN and TCN variants, especially RNN variants. Here we compare the proposed Multimodal Graph with RNN and TCN variants where the unimodal graph and GPFN in Multimodal Graph are replaced by them so as to investigate the effectiveness of graph neural networks on modeling sequence  (for Transformer \cite{transformer}, please refer to section~\ref{sec:multimodal} to see the comparison with Multimodal Transformer \cite{mult}).

From Table~\ref{t1}, it can be observed that GRU \cite{Cho2014Learning} and LSTM \cite{Hochreiter1997Long} perform slightly better than TCN \cite{TCN}, while our Multimodal Graph outperforms GRU and LSTM across the majority of the evaluation metrics. Specifically, our Multimodal Graph reaches the best performance on binary accuracy, F1 score, MAE, and Corr metrics, and ranks second on the 7-class accuracy. These results demonstrate Multimodal Graph's superiority in modeling sequential data. This is partly because our Multimodal Graph is able to automatically learn long-term dependency with a suitable adjacency matrix that links distant related nodes (time steps). Moreover, as indicated in section~\ref{sec:Analysis}, for each time step, Multimodal Graph can effectively link different related time steps with it, which is more interpretable, specific and representative. These results demonstrate GCN with an appropriate adjacency matrix as a novel and effective way of modelling sequences.

\textbf{Analysis of Training Time and Number of Parameters}: Intuitively, graph convolution dispenses with the recurrence nature and allows parallel operation in the time dimension, it is faster than RNN networks. To verify this point, we report the training time per batch on the CMU-MOSEI dataset, as shown in Table~\ref{t1}. It can be seen that under the same setting, training GCN network requires only 125ms per batch, compared to 933ms and 774ms per batch for GRU and LSTM, respectively. TCN network, as a recent competitive sequence learning model, takes 127ms per batch. This empirically demonstrates that GCN is much more efficient than RNNs and is comparable to TCN in terms of time complexity without sacrificing performance.
We also report the number of trainable parameters of the competing baselines to reflect space complexity. It can be seen that GRU and LSTM require 1,018,811 and 917,831 parameters respectively, both fewer than GCNs. This is because we only implement three layers of GRU/LSTM which performs best according to our experiment.
When we stack more layers, the performance decreases, which is reasonable because RNNs generally are more difficult to train when the network grows deeper. Additionally, TCN has 1,367,561 parameters, slightly more than our model.

\subsubsection{\textbf{Discussion of Adjacency Matrices}}\label{sec:adj_matrix}

\begin{table}[t]
\centering
 \caption{ \label{t2}\textbf{Discussion on the adjacency matrix on CMU-MOSEI.}}
\resizebox{.95\columnwidth}{!}{\begin{tabular}{c|c|c|c|c|c}
 \hline
    & Acc2 & Acc7 & F1 & MAE & Corr\\
     \hline
  All-one Matrix & 78.8 & 49.0 & 79.3 & 0.644 & 0.623\\
KNN & 80.3 & 45.8 & 80.6 & 0.659 & 0.625 \\
 GDM & 81.1 & 49.3 & 81.2 & 0.617 & 0.666 \\
Direct Learning & 80.7 & 49.3 & 81.2 & 0.618 & 0.659\\
 Indirect Learning & \textbf{81.4} & \textbf{49.7} &  \textbf{81.7} & \textbf{0.608} & \textbf{0.675}\\
 \hline
 \end{tabular}}
 \vspace{-0.3cm}
\end{table}%

To analyze the performance of different kinds of adjacency matrices, we conduct an experiment where different types of adjacency matrix are used in Multimodal Graph. To show the effectiveness of the proposed types of adjacency matrix, i.e., indirect learning, direct learning, GDM and KNN-based adjacency matrices, we additionally implement one comparative trial where the adjacency matrix is an all-one matrix which corresponds to a fully connected graph.
We can infer from Table~\ref{t2} that among all the adjacency matrices, the all-one matrix performs worst, which is understandable because the GCN in this case will reduce into a fully connected graph with no ability to discern various relatedness between nodes.
Besides, the KNN-based adjacency matrix reaches a relatively low performance, indicating that the Euclidean distance is not a perfect choice to determine the correlation between heterogeneous nodes.
In contrast, both the indirect learning method and the GDM achieve satisfactory results. This is because GDM is in line with the general pattern of sequence modelling which sticks out the present time step and dilutes the past ones.
And the direct learning method can learn such pattern in the absence of prior knowledge by gradient descent. Additionally, the indirect learning method, which is the default method to form an adjacency matrix, produces best results on all metrics. One possible reason is that it is both learnable and instance-specific, and therefore it is more representative and can be optimized to capture more subtle and complex relatedness among nodes.
To sum up, the comparative results that different types of adjacency matrices yield suggest that: (1) the proposed adjacency matrices are helpful in capturing useful information on the relatedness among nodes, compared to an all-one adjacency matrix that contains no such information, and (2) a learnable and meanwhile instance-specific adjacency matrix is crucial for capturing more relatedness information among nodes.


\begin{table}[t]
\centering
 \caption{ \label{t4}\textbf{ Performance of Multimodal Graph on CMU-MOSI.} The best performance is highlighted in bold.}
\resizebox{.98\columnwidth}{!}{\begin{tabular}{c|c|c|c|c|c}
 \hline
   Methods & Acc2 &  Acc7 & F1 & MAE & Corr\\
 \hline
 LF-LSTM & 74.5 & 31.3  & 74.3 & 1.042 & 0.608\\
 EF-LSTM & 73.7 & 32.2 & 73.5 & 1.038 & 0.594\\
  TFN \cite{Zadeh2017Tensor} & 77.9 & 32.4 & 75.0 & 1.040 & 0.616\\
  MFN \cite{Zadeh2018Memory} & 77.7 & 30.9 & 75.5 & 1.032 & 0.627\\
   MulT \cite{mult} & \textbf{80.6} & \textbf{35.3} & 79.3 & 0.972 & 0.681 \\
 \hline
   Multimodal Graph & \textbf{80.6} & 32.1 &  \textbf{80.5} & \textbf{0.933} & \textbf{0.684}\\
 \hline
 \end{tabular}}
\end{table}%

\begin{table}[t]
\centering
 \caption{ \label{t6}\textbf{ Performance of Multimodal Graph on CMU-MOSEI dataset.} The best performance is highlighted in bold.}
\resizebox{.98\columnwidth}{!}{\begin{tabular}{c|c|c|c|c|c}
 \hline
   Methods & Acc2 &  Acc7 & F1 & MAE & Corr\\
 \hline
 LF-LSTM & 79.5 & 48.0 & 79.6 & 0.632 & 0.650   \\
 EF-LSTM & 65.3 & 41.7 & 76.0 & 0.799 & 0.265  \\
 TFN \cite{Zadeh2017Tensor} & 79.5 & 49.3 & 78.9 & 0.613 & 0.673 \\
  MFN \cite{Zadeh2018Memory} & 80.6 & 49.1 & 80.0 & 0.612 & \textbf{0.687} \\
   MulT \cite{mult} & 80.1 & 49.0  & 80.9 & 0.630 & 0.664 \\
 \hline
   Multimodal Graph & \textbf{81.4} & \textbf{49.7} &  \textbf{81.7} & \textbf{0.608} & 0.675\\
 \hline
 \end{tabular}}
\end{table}%



\subsubsection{\textbf{Visualization of Adjacency Matrices}}
\begin{figure*}[htbp]
\centering
\subfigure[Indirect Learning (Language)]{
\begin{minipage}[t]{0.3\linewidth}
\centering
\includegraphics[width=2.2in]{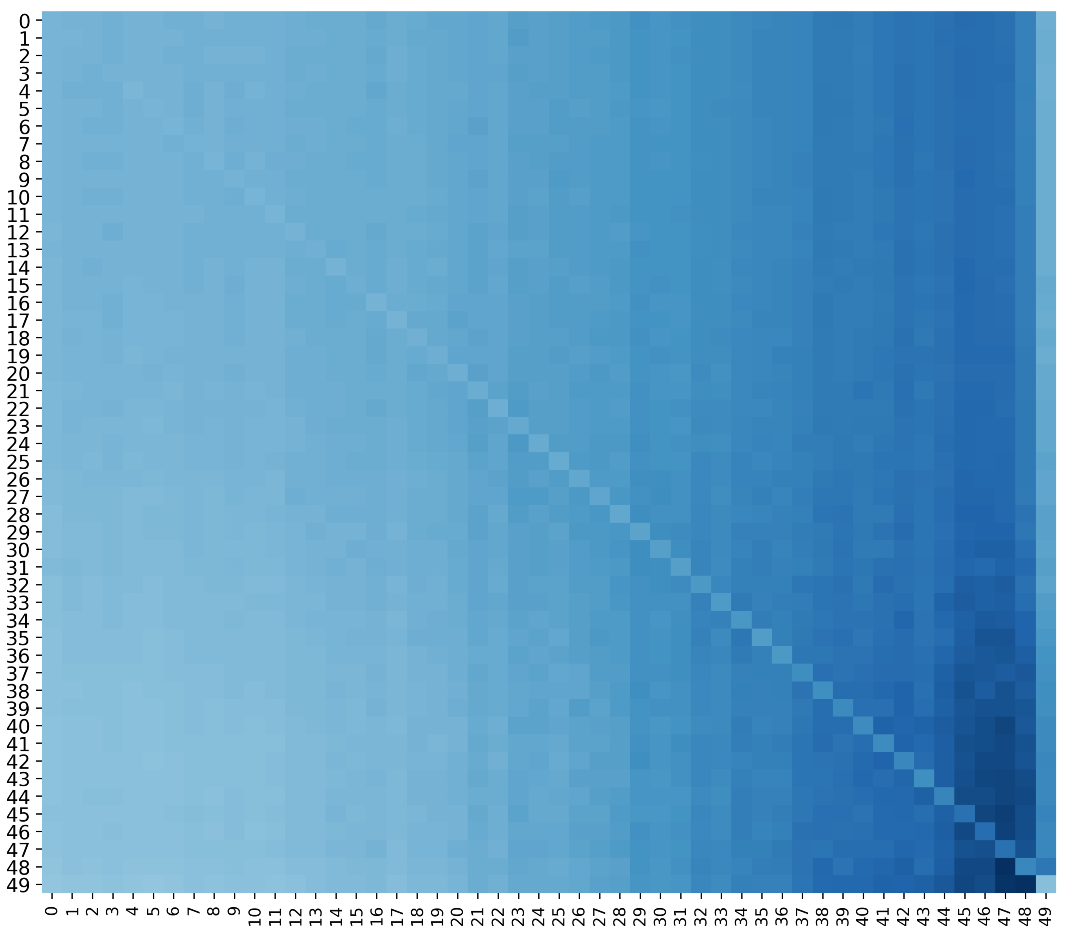}
\end{minipage}%
}
\subfigure[Indirect Learning (Acoustic)]{
\begin{minipage}[t]{0.3\linewidth}
\centering
\includegraphics[width=2.2in]{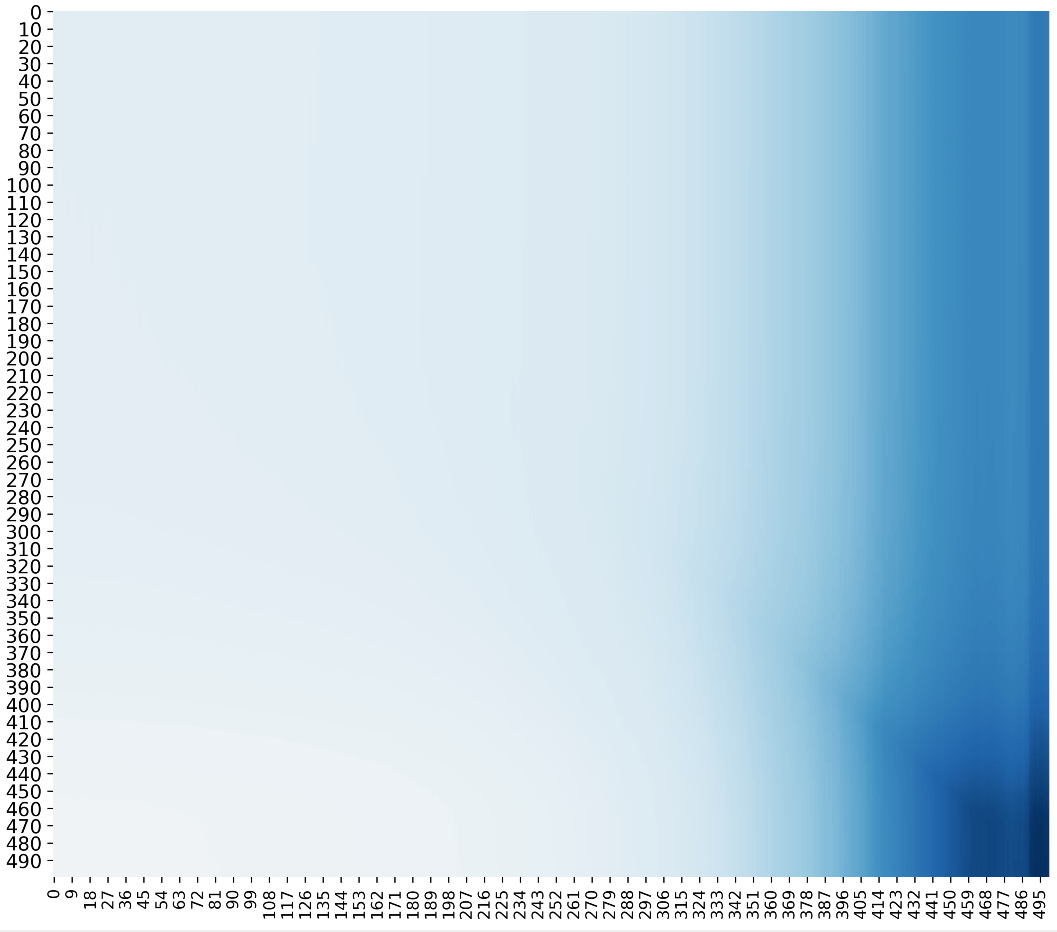}
\end{minipage}
}%
\subfigure[Indirect Learning (Visual)]{
\begin{minipage}[t]{0.3\linewidth}
\centering
\includegraphics[width=2.2in]{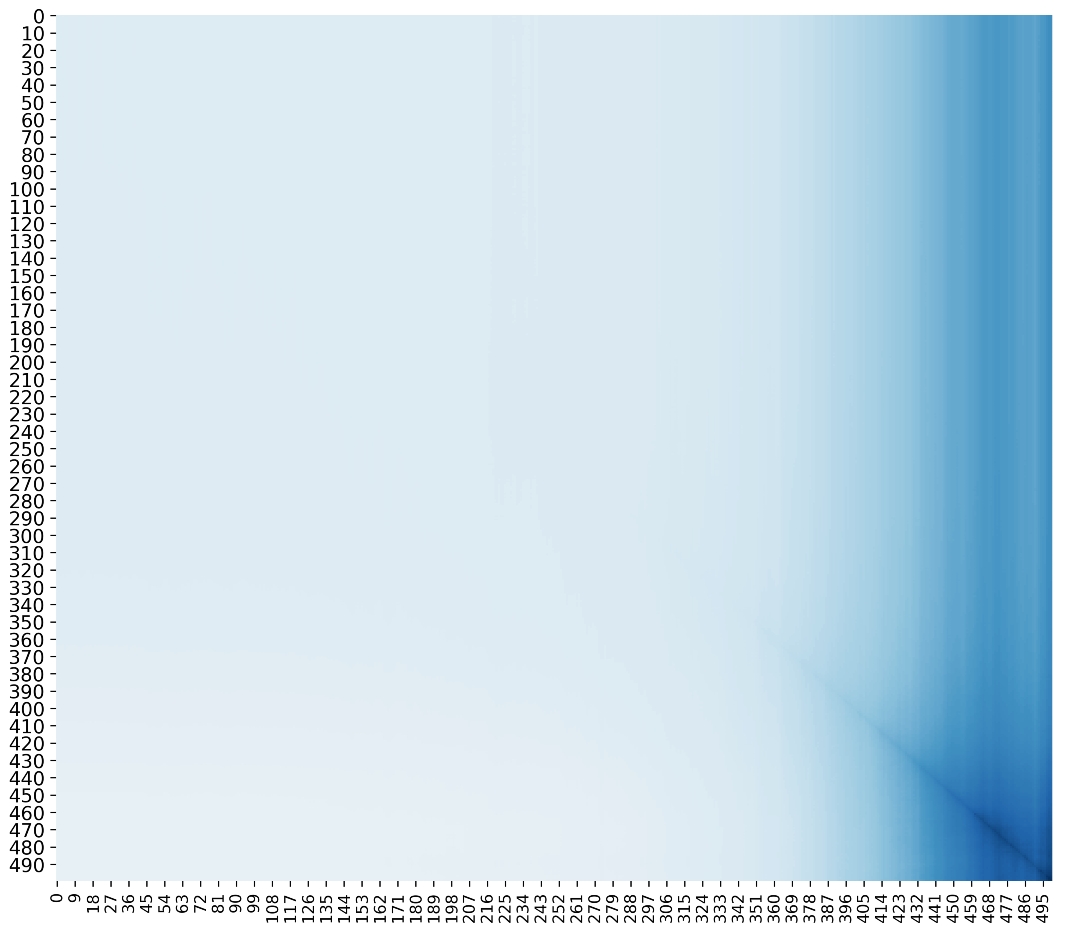}
\end{minipage}
}%
\centering
\caption{\label{7}\textbf{Visualization of Unimodal Indirect Learning Adjacency Matrices.} For indirect learning methods, since the adjacency matrices are instance-specific, we average the adjacency matrices of the testing instances and visualize the mean adjacency matrices. }
\end{figure*}

\begin{figure*}[htbp]
\centering
\subfigure[Indirect Learning (Cross-modal)]{
\begin{minipage}[t]{0.45\linewidth}
\centering
\includegraphics[width=3.3in]{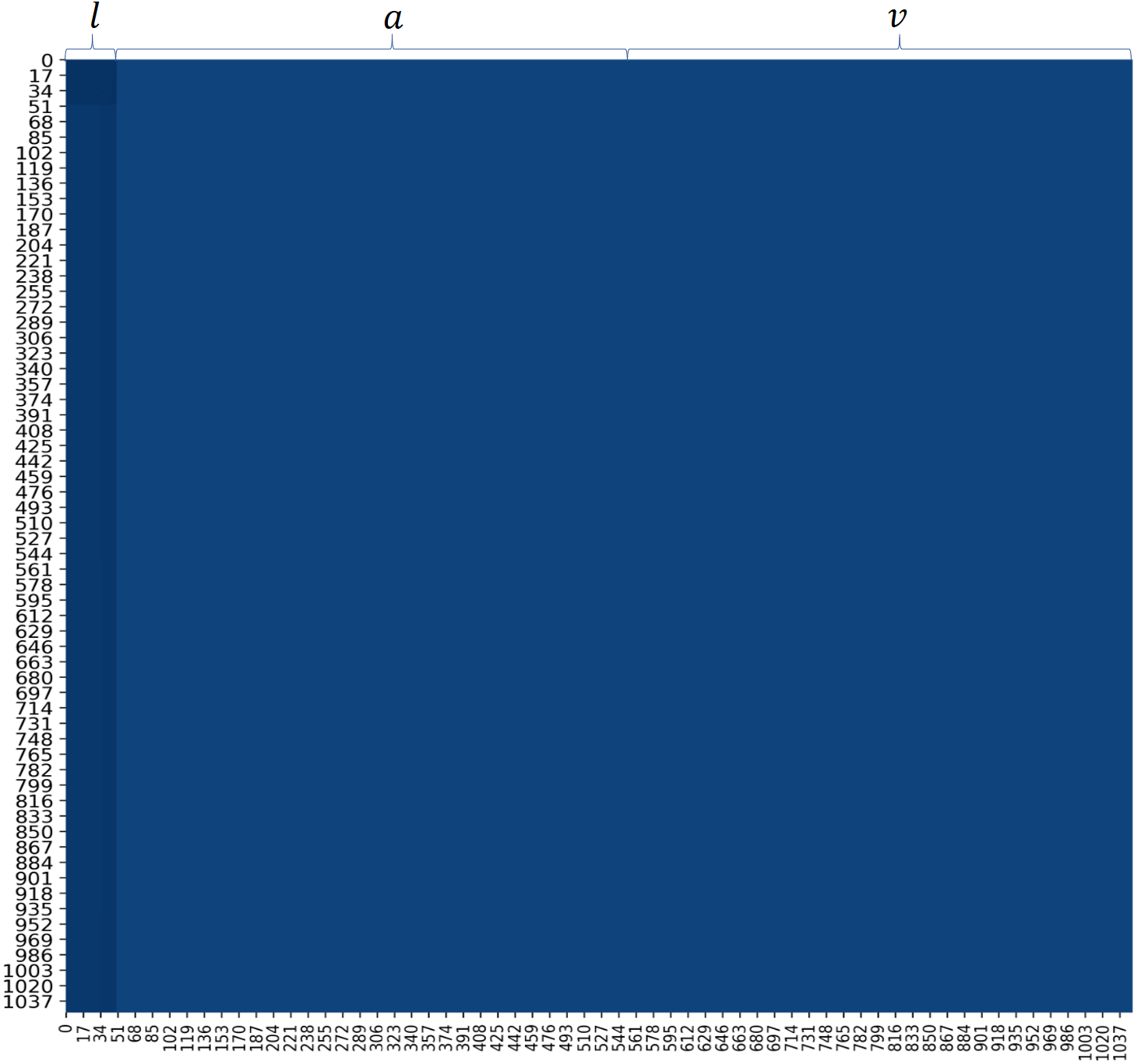}
\end{minipage}%
}
\subfigure[KNN (Cross-modal)]{
\begin{minipage}[t]{0.45\linewidth}
\centering
\includegraphics[width=3.3in]{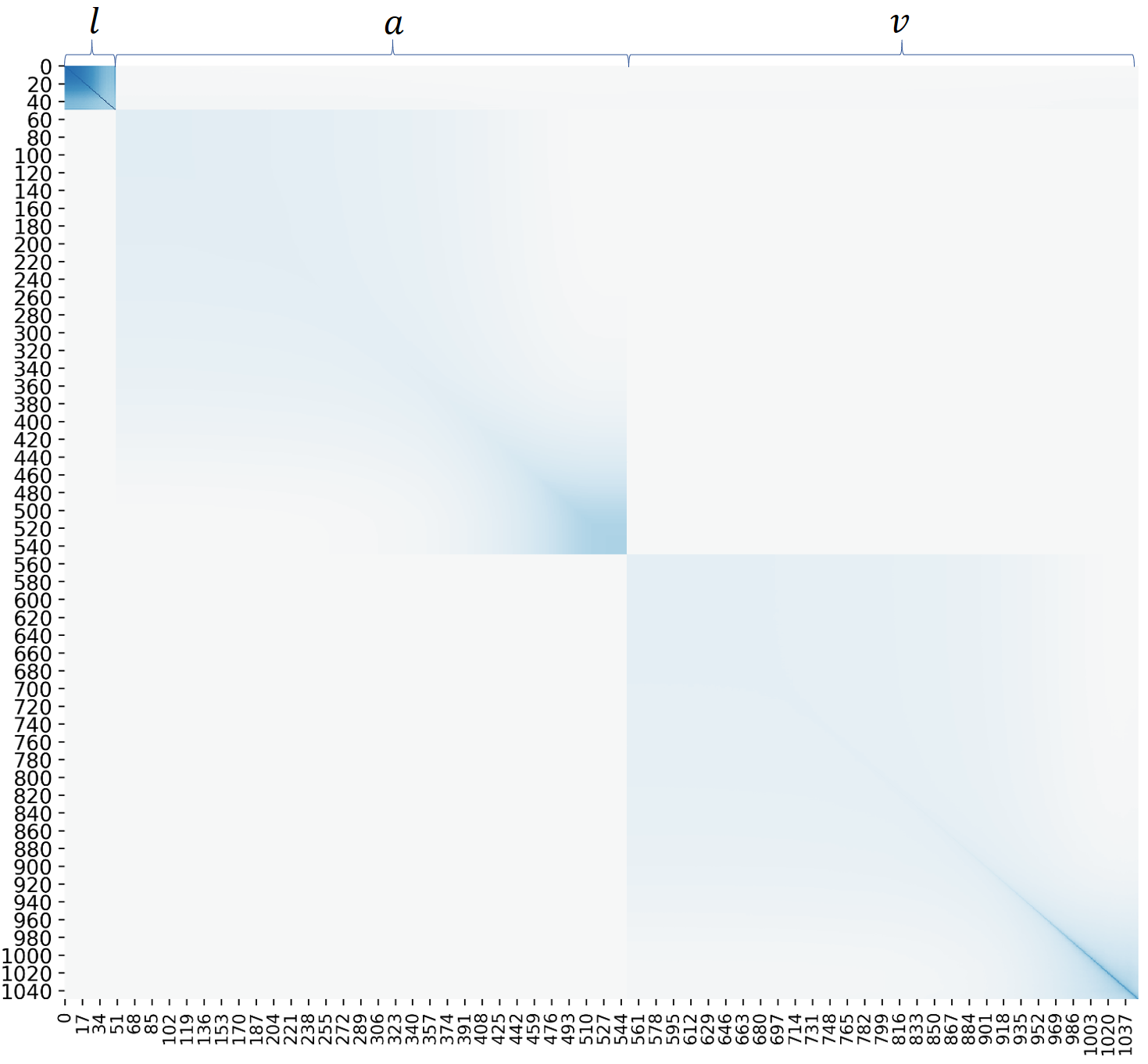}
\end{minipage}
}%
\centering
\caption{\label{8}\textbf{Visualization of Cross-modal Indirect Learning Adjacency Matrices.} Since the KNN-based and indirect learning methods are both instance-specific, we average the adjacency matrices of the testing instances and visualize the mean adjacency matrices. The first 50, the middle 500 and the latter 500 nodes belong to language, acoustic, and visual modality, respectively.}
\end{figure*}

To analyze the attributes of the indirect learning adjacency matrices, we provide a visualization of the unimodal and cross-modal indirect learning adjacency matrices. Instead of merely analyzing a few instances, we average the adjacency matrices of all the testing instances in the CMU-MOSEI dataset to reveal the general patterns of the adjacency matrices, which is more convincing and informative.

As can be inferred from Fig.~\ref{7}, for the visual and acoustic modalities, the latter portion of nodes have much more impact than their counterparts in that they have more intensive link association (the nodes tends to have more connections with the latter portion of nodes), indicting that the model heavily relies on the latter portion of nodes for prediction and they are more informative.
In contrast, the link association of different nodes in language modality is more even, indicting that the connections between different words are distributed evenly at nearly all temporal positions. Interestingly, the language adjacency matrix suggests that the language nodes have fewer self-connections. This is reasonable because during the graph convolution, we add self-loop operation such that the nodes can connect with themselves (see Eq.~\ref{eq1111}), so the adjacency matrix does not need to learn a strong self-connection value on the diagonal, otherwise it would be redundant.

We also conduct visualization of the cross-modal indirect learning adjacency matrix, and we provide the visualization of KNN-based cross-modal adjacency matrix for comparison. As can be inferred from Fig.~\ref{8}, for KNN-based method, the nodes from different modalities seem to have no direct and close interactions with each other, and these nodes only interact with their neighboring nodes that come from the same modality. This means that by using the KNN-based adjacency matrix, the cross-modal interactions cannot be effectively explored, which may explain why it performs worse than other methods. The visualization of the KNN-based adjacency matrix also suggests that the distribution gap between different modalities actually exists and we need to handle it during modality fusion \cite{ARGF}. In contrast, for adjacency matrix of the indirect learning method, interestingly, the mean cross-modal adjacency matrix suggests that the nodes tend to connect with the language nodes more closely, which indicates that the language nodes are more important and informative. This scenario can be partly explained by the fact that language is more important than the other modalities, as revealed in \cite{MCTN,HFFN}. Additionally, our visualization suggests the language nodes have the strongest connection with each other in the cross-modal adjacency matrix.

\subsubsection{\textbf{Comparison with Baselines}}\label{sec:multimodal}
We compare our Multimodal Graph with the competing baselines on two benchmark datasets CMU-MOSI \cite{Zadeh2016Multimodal} and CMU-MOSEI \cite{mosei}, and the results are presented in Table \ref{t4} and Table \ref{t6}, respectively. It can be seen that Multimodal Graph yields best results on most of the metrics on two datasets. Specifically, our model surpasses the state-of-the-art unaligned cross-modal fusion method MulT \cite{mult} in terms of all metrics, except 7-class accuracy on CMU-MOSI, which demonstrates the effectiveness of graph convolution and graph pooling in learning sequential data, compared to Transformer which is employed in MulT.


\subsubsection{\textbf{Analysis of Model Complexity}}

\begin{table}[t]
\centering
 \caption{ \label{t222}The Comparison of Model Complexity on CMU-MOSEI.}
 \resizebox{.9\columnwidth}{!}{\begin{tabular}{c|c}
 \hline
   Methods  & Number of Parameters \\
 \hline
 TFN \cite{Zadeh2017Tensor}  & 1,002,471  \\
  MulT \cite{mult}  & 1,901,161  \\
 MFN \cite{Zadeh2018Memory}  & \textbf{963,777}  \\
 \hline
 Multimodal Graph  & 1,225,400 \\
 \hline
 \end{tabular}}
\end{table}%

To analyze the model complexity of Multimodal Graph, we use the number of trainable parameters on the CMU-MOSEI dataset as the proxy for its space complexity, and compare it to the state-of-the-art unaligned fusion methods, as reported in Table \ref{t222}. It can be seen that our Multimodal Graph requires 1,225,400 trainable parameters on CMU-MOSEI, which is 64.46\% of the number of parameters of MulT. Although Multimodal Graph requires fewer trainable parameters than the current state-of-the-art model MulT, it still outperforms MulT in practice as shown in the previous subsection. This advantage in space complexity is significant because it shows that in addition to being effective in modeling sequential data, our graph convolution network is also more efficient than the variant of Transformer, which is one dominant sequence learning method. This further validates graph neural network as a novel means of learning sequential data. To sum up, given the high empirical performance of Multimodal Graph, the space complexity of Multimodal Graph is moderate compared to state-of-the-art unaligned fusion methods.


\subsubsection{\textbf{Discussion of Different Graph Neural Networks}}

\begin{table}[t]
\centering
 \caption{ \label{t7}\textbf{ Discussion on the Concrete Graph Neural Networks on CMU-MOSEI.} For the case of `GAT' and `GIN', we replace the defaulted graph convolution model GraphSAGE in unimodal graphs and GPFN with the corresponding GAT and GIN model. For the case of `DiffPool', we replace the max/mean grpah pooling and link similarity pooling in GPFN with DiffPool.}
\resizebox{.95\columnwidth}{!}{\begin{tabular}{c|c|c|c|c|c}
 \hline
  Models  & Acc2 &   Acc7 & F1 & MAE & Corr\\
 \hline
 GAT \cite{GAT} & 80.3 & 49.0 & 80.4  & 0.627 & 0.650\\
 GIN \cite{DIN} & 81.1 & 49.0 & 81.6 & 0.615 & \textbf{0.676} \\
 GraphSAGE \cite{GraphSAGE} & \textbf{81.4} & \textbf{49.7} &  \textbf{81.7} & \textbf{0.608} & 0.675\\
 \hline
  DiffPool \cite{diffpool} & 81.1 & \textbf{49.8} & 81.3 & 0.611 & 0.670\\
  GPFN  & \textbf{81.4} & 49.7 &  \textbf{81.7} & \textbf{0.608} & \textbf{0.675}\\
 \hline
 \end{tabular}}
 \vspace{-0.3cm}
\end{table}%

Since our proposed Multimodal Graph is independent of the concrete GCN structure, we can choose any GCN structure to implement our model. This subsection compares different current GCN structures to analyze which kind of GCN structures performs best. Specifically, we compare the defaulted GraphSAGE \cite{GraphSAGE} with GAT \cite{GAT} and GIN \cite{DIN}. From Table~\ref{t7} it can be seen that GraphSAGE \cite{GraphSAGE} reaches the best performance among all the compared GCNs. In addition, the results of GIN \cite{DIN} and GAT \cite{GAT} are also satisfactory, indicating the generalization ability of Multimodal Graph.

As for the graph pooling methods, we implement and evaluate DiffPool \cite{diffpool} as the baseline to compare with the proposed GPFN. DiffPool \cite{diffpool} achieves a relatively good result and is still inferior to our model. To be more specific, DiffPool outperforms our GPFN in terms of 7-class accuracy by 0.1 points, while our model outperforms DiffPool on the rest of the evaluation metrics. These results demonstrate the effectiveness of our graph pooling method.

\section{Conclusion}\label{sec:Conclusion}
In this paper, we focus on the task of multimodal sentiment analysis, which is typically a sequential learning task. We seek to employ the popular graph neural networks to process multimodal sequences, which is free of the recurrent structure and proves more efficient and more competitive. Specifically, we develop a unimodal graph for each single modality to explore their intra-modal dynamics, and a graph pooling fusion network over the unimodal graphs to learn inter-modal dynamics. We define and compare multiple ways to construct an adjacency matrix for a sequence, showing that a learnable and instance-specific method performs best. Extensive experiments comparing the proposed GCN-based model with RNN and TCN variants also reveal the superiority of GCN in terms of efficiency and performance. However, Our Multimodal Graph is non-causal, not applicable to some language processing tasks. In the future, we aim to develop a causal GCN-based model to process sequence.


\ifCLASSOPTIONcaptionsoff
  \newpage
\fi

\bibliographystyle{IEEEtran}


%

%

\bibliography{sentiment2}

\begin{thebibliography}{10}
\providecommand{\url}[1]{#1}
\csname url@samestyle\endcsname
\providecommand{\newblock}{\relax}
\providecommand{\bibinfo}[2]{#2}
\providecommand{\BIBentrySTDinterwordspacing}{\spaceskip=0pt\relax}
\providecommand{\BIBentryALTinterwordstretchfactor}{4}
\providecommand{\BIBentryALTinterwordspacing}{\spaceskip=\fontdimen2\font plus
\BIBentryALTinterwordstretchfactor\fontdimen3\font minus
  \fontdimen4\font\relax}
\providecommand{\BIBforeignlanguage}[2]{{%
\expandafter\ifx\csname l@#1\endcsname\relax
\typeout{** WARNING: IEEEtran.bst: No hyphenation pattern has been}%
\typeout{** loaded for the language `#1'. Using the pattern for}%
\typeout{** the default language instead.}%
\else
\language=\csname l@#1\endcsname
\fi
#2}}
\providecommand{\BIBdecl}{\relax}
\BIBdecl

\bibitem{RMFN}
P.~P. Liang, Z.~Liu, A.~Zadeh, and L.~P. Morency, ``Multimodal language
  analysis with recurrent multistage fusion,'' in \emph{EMNLP}, 2018, pp.
  150--161.

\bibitem{Zadeh2017Tensor}
A.~Zadeh, M.~Chen, S.~Poria, E.~Cambria, and L.~P. Morency, ``Tensor fusion
  network for multimodal sentiment analysis,'' in \emph{EMNLP}, 2017, pp.
  1114--1125.

\bibitem{RNN}
M.~W. Goudreau, C.~L. Giles, S.~T. Chakradhar, and .~Chen, D., ``First-order
  versus second-order single-layer recurrent neural networks,'' \emph{IEEE
  Transactions on Neural Networks}, vol.~5, no.~3, pp. 511--513, 1994.

\bibitem{Hochreiter1997Long}
S.~Hochreiter and J.~Schmidhuber, ``Long short-term memory,'' \emph{Neural
  Computation}, vol.~9, no.~8, pp. 1735--1780, 1997.

\bibitem{Cho2014Learning}
K.~Cho, B.~Van~Merrienboer, C.~Gulcehre, D.~Bahdanau, F.~Bougares, H.~Schwenk,
  and Y.~Bengio, ``Learning phrase representations using rnn encoder-decoder
  for statistical machine translation,'' in \emph{EMNLP}, 2014, pp. 1724--1734.

\bibitem{Multilogue-Net}
A.~Shenoy and A.~Sardana, ``Multilogue-net: A context aware rnn for multi-modal
  emotion detection and sentiment analysis in conversation,'' \emph{arXiv
  preprint arXiv:2002.08267}, 2020.

\bibitem{RAVEN}
Y.~Wang, Y.~Shen, Z.~Liu, P.~P. Liang, A.~Zadeh, and L.-P. Morency, ``Words can
  shift: Dynamically adjusting word representations using nonverbal
  behaviors,'' in \emph{Proceedings of the AAAI Conference on Artificial
  Intelligence}, vol.~33, 2019, pp. 7216--7223.

\bibitem{Zadeh2018Memory}
A.~Zadeh, P.~P. Liang, N.~Mazumder, S.~Poria, E.~Cambria, and L.~P. Morency,
  ``Memory fusion network for multi-view sequential learning,'' in \emph{AAAI},
  2018, pp. 5634--5641.

\bibitem{Bengio1994Learning}
Y.~Bengio, P.~Simard, and P.~Frasconi, ``Learning long-term dependencies with
  gradient descent is difficult,'' \emph{IEEE Transactions on Neural Networks},
  vol.~5, no.~2, pp. 157--166, 1994.

\bibitem{MFRM}
S.~{Mai}, H.~{Hu}, J.~{Xu}, and S.~{Xing}, ``Multi-fusion residual memory
  network for multimodal human sentiment comprehension,'' \emph{IEEE
  Transactions on Affective Computing}, pp. 1--1, 2020.

\bibitem{mosei}
A.~Zadeh, P.~P. Liang, J.~Vanbriesen, S.~Poria, E.~Tong, E.~Cambria, M.~Chen,
  and L.~P. Morency, ``Multimodal language analysis in the wild: Cmu-mosei
  dataset and interpretable dynamic fusion graph,'' in \emph{ACL}, 2018, pp.
  2236--2246.

\bibitem{P2FA}
J.~Yuan and M.~Liberman, ``{Speaker identification on the SCOTUS corpus},''
  \emph{Acoustical Society of America Journal}, vol. 123, p. 3878, 2008.

\bibitem{DIN}
K.~Xu, W.~Hu, J.~Leskovec, and S.~Jegelka, ``How powerful are graph neural
  networks?'' in \emph{ICLR}, 2019.

\bibitem{Micheli2009Neural}
A.~Micheli, ``Neural network for graphs: A contextual constructive approach,''
  \emph{IEEE Transactions on Neural Networks}, vol.~20, no.~3, pp. p.498--511,
  2009.

\bibitem{4700287}
F.~{Scarselli}, M.~{Gori}, A.~C. {Tsoi}, M.~{Hagenbuchner}, and
  G.~{Monfardini}, ``The graph neural network model,'' \emph{IEEE Transactions
  on Neural Networks}, vol.~20, no.~1, pp. 61--80, 2009.

\bibitem{TCN}
\BIBentryALTinterwordspacing
S.~Bai, J.~Z. Kolter, and V.~Koltun, ``An empirical evaluation of generic
  convolutional and recurrent networks for sequence modeling,'' \emph{Arxiv
  preprint Arxiv: 1803.01271}, 2018. [Online]. Available:
  \url{http://arxiv.org/abs/1803.01271}
\BIBentrySTDinterwordspacing

\bibitem{Trellis}
S.~Bai, J.~Kolter, and V.~Koltun, ``Trellis networks for sequence modeling,''
  in \emph{ICLR}, 2019.

\bibitem{GraphSAGE}
W.~Hamilton, Z.~Ying, and J.~Leskovec, ``Inductive representation learning on
  large graphs,'' in \emph{Advances in neural information processing systems},
  2017, pp. 1024--1034.

\bibitem{8269806}
T.~Baltru\v{s}aitis, C.~Ahuja, and L.-P. Morency, ``Multimodal machine
  learning: A survey and taxonomy,'' \emph{IEEE Transactions on Pattern
  Analysis and Machine Intelligence}, vol.~41, no.~2, pp. 423--443, Feb 2019.

\bibitem{Multi-view}
G.~Aguilar, V.~Rozgic, W.~Wang, and C.~Wang, ``Multimodal and multi-view models
  for emotion recognition,'' in \emph{Proceedings of the 57th Annual Meeting of
  the Association for Computational Linguistics}, 2019, pp. 991--1002.

\bibitem{Xu2013A}
S.~Sun, ``A survey of multi-view machine learning,'' \emph{Neural Computing and
  Applications}, vol.~23, no. 7-8, pp. 2031--2038, 2013.

\bibitem{Wollmer2013YouTube}
M.~Wollmer, F.~Weninger, T.~Knaup, B.~Schuller, C.~Sun, K.~Sagae, and L.~P.
  Morency, ``Youtube movie reviews: Sentiment analysis in an audio-visual
  context,'' \emph{IEEE Intelligent Systems}, vol.~28, no.~3, pp. 46--53, 2013.

\bibitem{Rozgic2012Ensemble}
V.~Rozgic, S.~Ananthakrishnan, S.~Saleem, R.~Kumar, and R.~Prasad, ``Ensemble
  of svm trees for multimodal emotion recognition,'' in \emph{Signal and
  Information Processing Association Summit and Conference}, 2012, pp. 1--4.

\bibitem{Poria2017Convolutional}
S.~Poria, I.~Chaturvedi, E.~Cambria, and A.~Hussain, ``Convolutional mkl based
  multimodal emotion recognition and sentiment analysis,'' in \emph{Proceedings
  of IEEE International Conference on Data Mining (ICDM)}, 2016, pp. 439--448.

\bibitem{Poria2017Context}
S.~Poria, E.~Cambria, D.~Hazarika, N.~Majumder, A.~Zadeh, and L.~P. Morency,
  ``Context-dependent sentiment analysis in user-generated videos,'' in
  \emph{ACL}, 2017, pp. 873--883.

\bibitem{Wu2010Emotion}
C.~H. Wu and W.~B. Liang, ``Emotion recognition of affective speech based on
  multiple classifiers using acoustic-prosodic information and semantic
  labels,'' \emph{IEEE Transactions on Affective Computing}, vol.~2, no.~1, pp.
  10--21, 2010.

\bibitem{Nojavanasghari2016Deep}
B.~Nojavanasghari, D.~Gopinath, J.~Koushik, and L.~P. Morency, ``Deep
  multimodal fusion for persuasiveness prediction,'' in \emph{Proceedings of
  ACM International Conference on Multimodal Interaction}, 2016, pp. 284--288.

\bibitem{Personality}
\BIBentryALTinterwordspacing
O.~Kampman, E.~J. Barezi, D.~Bertero, and P.~Fung, ``Investigating audio,
  visual, and text fusion methods for end-to-end automatic personality
  prediction,'' in \emph{ACL short paper}, 2018. [Online]. Available:
  \url{http://arxiv.org/abs/1805.00705}
\BIBentrySTDinterwordspacing

\bibitem{Zadeh2016MOSI}
A.~Zadeh, R.~Zellers, E.~Pincus, and L.~P. Morency, ``Mosi: Multimodal corpus
  of sentiment intensity and subjectivity analysis in online opinion videos,''
  \emph{IEEE Intelligent Systems}, vol.~31, no.~6, pp. 82--88, 2016.

\bibitem{T2FN}
\BIBentryALTinterwordspacing
P.~P. Liang, Z.~Liu, Y.-H.~H. Tsai, Q.~Zhao, R.~Salakhutdinov, and L.-P.
  Morency, ``Learning representations from imperfect time series data via
  tensor rank regularization,'' in \emph{Proceedings of the 57th Annual Meeting
  of the Association for Computational Linguistics}.\hskip 1em plus 0.5em minus
  0.4em\relax Florence, Italy: Association for Computational Linguistics, Jul.
  2019, pp. 1569--1576. [Online]. Available:
  \url{https://www.aclweb.org/anthology/P19-1152}
\BIBentrySTDinterwordspacing

\bibitem{Liu2018Efficient}
Z.~Liu, Y.~Shen, P.~P. Liang, A.~Zadeh, and L.~P. Morency, ``Efficient low-rank
  multimodal fusion with modality-specific factors,'' in \emph{ACL}, 2018, pp.
  2247--2256.

\bibitem{LMFN}
S.~{Mai}, S.~{Xing}, and H.~{Hu}, ``Locally confined modality fusion network
  with a global perspective for multimodal human affective computing,''
  \emph{IEEE Transactions on Multimedia}, vol.~22, no.~1, pp. 122--137, 2020.

\bibitem{HFFN}
\BIBentryALTinterwordspacing
S.~Mai, H.~Hu, and S.~Xing, ``Divide, conquer and combine: Hierarchical feature
  fusion network with local and global perspectives for multimodal affective
  computing,'' in \emph{Proceedings of the 57th Conference of the Association
  for Computational Linguistics}.\hskip 1em plus 0.5em minus 0.4em\relax
  Florence, Italy: Association for Computational Linguistics, Jul. 2019, pp.
  481--492. [Online]. Available:
  \url{https://www.aclweb.org/anthology/P19-1046}
\BIBentrySTDinterwordspacing

\bibitem{HPFN}
M.~Hou, J.~Tang, J.~Zhang, W.~Kong, and Q.~Zhao, ``Deep multimodal multilinear
  fusion with high-order polynomial pooling,'' in \emph{Advances in Neural
  Information Processing Systems}, 2019, pp. 12\,113--12\,122.

\bibitem{MCTN}
H.~Pham, P.~P. Liang, T.~Manzini, L.~P. Morency, and P.~Barnab\v{a}s, ``Found
  in translation: Learning robust joint representations by cyclic translations
  between modalities,'' in \emph{AAAI}, 2019, pp. 6892--6899.

\bibitem{mult}
\BIBentryALTinterwordspacing
Y.-H.~H. Tsai, S.~Bai, P.~P. Liang, J.~Z. Kolter, L.-P. Morency, and
  R.~Salakhutdinov, ``Multimodal transformer for unaligned multimodal language
  sequences,'' in \emph{Proceedings of the 57th Annual Meeting of the
  Association for Computational Linguistics}.\hskip 1em plus 0.5em minus
  0.4em\relax Florence, Italy: Association for Computational Linguistics, Jul.
  2019, pp. 6558--6569. [Online]. Available:
  \url{https://www.aclweb.org/anthology/P19-1656}
\BIBentrySTDinterwordspacing

\bibitem{ARGF}
S.~Mai, H.~Hu, and S.~Xing, ``Modality to modality translation: An adversarial
  representation learning and graph fusion network for multimodal fusion,'' in
  \emph{Proceedings of the AAAI Conference on Artificial Intelligence},
  vol.~34, no.~01, 2020, pp. 164--172.

\bibitem{MFM}
Y.~H.~H. Tsai, P.~P. Liang, A.~Zadeh, L.~P. Morency, and R.~Salakhutdinov,
  ``Learning factorized multimodal representations,'' in \emph{ICLR}, 2019.

\bibitem{MMB}
P.~P. Liang, Y.~C. Lim, Y.~H. Tsai, R.~R. Salakhutdinov, and L.-P. Morency,
  ``Strong and simple baselines for multimodal utterance embeddings,'' in
  \emph{NAACL}, 2019, pp. 2599--2609.

\bibitem{CIA}
D.~S. Chauhan, M.~S. Akhtar, A.~Ekbal, and P.~Bhattacharyya, ``Context-aware
  interactive attention for multi-modal sentiment and emotion analysis,'' in
  \emph{Proceedings of the 2019 Conference on Empirical Methods in Natural
  Language Processing and the 9th International Joint Conference on Natural
  Language Processing (EMNLP-IJCNLP)}, 2019, pp. 5651--5661.

\bibitem{MTL}
M.~S. Akhtar, D.~Chauhan, D.~Ghosal, S.~Poria, A.~Ekbal, and P.~Bhattacharyya,
  ``Multi-task learning for multi-modal emotion recognition and sentiment
  analysis,'' in \emph{Proceedings of the 2019 Conference of the North American
  Chapter of the Association for Computational Linguistics: Human Language
  Technologies, Volume 1 (Long and Short Papers)}, 2019, pp. 370--379.

\bibitem{Chen2018Multimodal}
M.~Chen, S.~Wang, P.~P. Liang, T.~Baltrus$\check{a}$itis, A.~Zadeh, and L.~P.
  Morency, ``Multimodal sentiment analysis with word-level fusion and
  reinforcement learning,'' in \emph{19th ACM International Conference on
  Multimodal Interaction (ICMI'17)}, 2017, pp. 163--171.

\bibitem{Gu2018Multimodal}
Y.~Gu, K.~Yang, S.~Fu, S.~Chen, X.~Li, and I.~Marsic, ``Multimodal affective
  analysis using hierarchical attention strategy with word-level alignment,''
  in \emph{ACL}, 2018, pp. 2225--2235.

\bibitem{DHF}
E.~Georgiou, C.~Papaioannou, and A.~Potamianos, ``Deep hierarchical fusion with
  application in sentiment analysis,'' \emph{Proc. Interspeech 2019}, pp.
  1646--1650, 2019.

\bibitem{Zadeh2018Multi}
A.~Zadeh, P.~P. Liang, S.~Poria, P.~Vij, E.~Cambria, and L.~P. Morency,
  ``Multi-attention recurrent network for human communication comprehension,''
  in \emph{AAAI}, 2018, pp. 5642--5649.

\bibitem{GCN}
T.~N. Kipf and M.~Welling, ``Semi-supervised classification with graph
  convolutional networks,'' in \emph{ICLR}, 2016.

\bibitem{SEAL}
M.~Zhang and Y.~Chen, ``Link prediction based on graph neural networks,'' in
  \emph{Advances in Neural Information Processing Systems}, 2018, pp.
  5165--5175.

\bibitem{DGCNN}
M.~Zhang, Z.~Cui, M.~Neumann, and Y.~Chen, ``An end-to-end deep learning
  architecture for graph classification,'' in \emph{Thirty-Second AAAI
  Conference on Artificial Intelligence}, 2018.

\bibitem{DBLP:journals/corr/abs-1903-11960}
\BIBentryALTinterwordspacing
L.~Franceschi, M.~Niepert, M.~Pontil, and X.~He, ``Learning discrete structures
  for graph neural networks,'' in \emph{ICML}, 2019. [Online]. Available:
  \url{http://arxiv.org/abs/1903.11960}
\BIBentrySTDinterwordspacing

\bibitem{diffpool}
Z.~Ying, J.~You, C.~Morris, X.~Ren, W.~Hamilton, and J.~Leskovec,
  ``Hierarchical graph representation learning with differentiable pooling,''
  in \emph{Advances in neural information processing systems}, 2018, pp.
  4800--4810.

\bibitem{Yuan2020StructPoolSG}
H.~Yuan and S.~Ji, ``Structpool: Structured graph pooling via conditional
  random fields,'' in \emph{ICLR}, 2020.

\bibitem{Olson1977From}
D.~Olson, ``From utterance to text: The bias of language in speech and
  writing,'' \emph{Harvard Educational Review}, vol.~47, no.~3, pp. 257--281,
  1977.

\bibitem{transformer}
A.~Vaswani, N.~Shazeer, N.~Parmar, J.~Uszkoreit, L.~Jones, A.~N. Gomez,
  {\L}.~Kaiser, and I.~Polosukhin, ``Attention is all you need,'' in
  \emph{Advances in neural information processing systems}, 2017, pp.
  5998--6008.

\bibitem{GAT}
P.~Veli{\v{c}}kovi{\'c}, G.~Cucurull, A.~Casanova, A.~Romero, P.~Lio, and
  Y.~Bengio, ``Graph attention networks,'' in \emph{ICLR}, 2018.

\bibitem{TCNN}
A.~Pandey and D.~Wang, ``Tcnn: Temporal convolutional neural network for
  real-time speech enhancement in the time domain,'' in \emph{ICASSP 2019-2019
  IEEE International Conference on Acoustics, Speech and Signal Processing
  (ICASSP)}.\hskip 1em plus 0.5em minus 0.4em\relax IEEE, 2019, pp. 6875--6879.

\bibitem{pmlr-v70-gehring17a}
\BIBentryALTinterwordspacing
J.~Gehring, M.~Auli, D.~Grangier, D.~Yarats, and Y.~N. Dauphin, ``Convolutional
  sequence to sequence learning,'' in \emph{Proceedings of the 34th
  International Conference on Machine Learning}, ser. Proceedings of Machine
  Learning Research, D.~Precup and Y.~W. Teh, Eds., vol.~70.\hskip 1em plus
  0.5em minus 0.4em\relax International Convention Centre, Sydney, Australia:
  PMLR, 06--11 Aug 2017, pp. 1243--1252. [Online]. Available:
  \url{http://proceedings.mlr.press/v70/gehring17a.html}
\BIBentrySTDinterwordspacing

\bibitem{Zadeh2016Multimodal}
A.~Zadeh, R.~Zellers, E.~Pincus, and L.~P. Morency, ``Multimodal sentiment
  intensity analysis in videos: Facial gestures and verbal messages,''
  \emph{IEEE Intelligent Systems}, vol.~31, no.~6, pp. 82--88, 11 2016.

\bibitem{CTC}
A.~Graves, S.~Fern{\'a}ndez, F.~Gomez, and J.~Schmidhuber, ``Connectionist
  temporal classification: labelling unsegmented sequence data with recurrent
  neural networks,'' in \emph{Proceedings of the 23rd international conference
  on Machine learning}, 2006, pp. 369--376.

\bibitem{Kingma2014Adam}
D.~P. Kingma and J.~Ba, ``Adam: A method for stochastic optimization,'' in
  \emph{Proceedings of International Conference on Learning Representations
  (ICLR)}, 2015.

\bibitem{Gkoumas2021WhatMT}
D.~Gkoumas, Q.~Li, C.~Lioma, Y.~Yu, and D.~wei Song, ``What makes the
  difference? an empirical comparison of fusion strategies for multimodal
  language analysis,'' \emph{Information Fusion}, vol.~66, pp. 184--197, 2021.

\bibitem{pennington2014glove}
\BIBentryALTinterwordspacing
J.~Pennington, R.~Socher, and C.~D. Manning, ``Glove: Global vectors for word
  representation,'' in \emph{Empirical Methods in Natural Language Processing
  (EMNLP)}, 2014, pp. 1532--1543. [Online]. Available:
  \url{http://www.aclweb.org/anthology/D14-1162}
\BIBentrySTDinterwordspacing

\bibitem{Degottex2014COVAREP}
G.~Degottex, J.~Kane, T.~Drugman, T.~Raitio, and S.~Scherer, ``Covarep: A
  collaborative voice analysis repository for speech technologies,'' in
  \emph{ICASSP}, 2014, pp. 960--964.

\end{thebibliography}

\end{document}